\documentclass{article}

\usepackage{arxiv}

\usepackage[utf8]{inputenc} 
\usepackage[T1]{fontenc}    
\usepackage{hyperref}       
\usepackage{url}            
\usepackage{booktabs}       
\usepackage{amsfonts}       
\usepackage{amsmath}        
\usepackage{amssymb}        
\usepackage{nicefrac}       
\usepackage{microtype}      
\usepackage{graphicx}
\usepackage{array}          
\usepackage{enumitem}       
\usepackage{algorithm}      
\usepackage{algpseudocode}
\usepackage[numbers]{natbib} 
\usepackage{doi}
\usepackage{orcidlink}

\title{Calibrated Multichannel Monocular Ranging From Standardized License Plates With Metrology-Exact Validation}

\author{Manognya Lokesh Reddy\\
	Department of Computer and Information Science\\
	University of Michigan\\
	Dearborn, MI 48128 \\
	\texttt{manognya@umich.edu} \\
	\And
	Zheng Liu\thanks{Corresponding author.} \orcidlink{0000-0003-4869-8893}\\
	Department of Industry and Manufacturing Systems Engineering\\
	University of Michigan\\
	Dearborn, MI 48128 \\
	\texttt{zhengtl@umich.edu} \\
}

\renewcommand{\shorttitle}{Calibrated Multichannel Monocular Plate Ranging}

\hypersetup{
pdftitle={Calibrated Multichannel Monocular Ranging From Standardized License Plates With Metrology-Exact Validation},
pdfsubject={License-plate monocular ranging: system improvements and simulation validation},
pdfauthor={Manognya Lokesh Reddy, Zheng Liu},
pdfkeywords={monocular ranging, license plate, inverse-variance fusion, simulation validation, forward collision warning},
}

\begin{document}
\maketitle

\begin{abstract}
	Longitudinal driver assistance depends on the distance to the vehicle ahead, a quantity normally supplied by radar, laser scanner, or stereo pair. However, a low-cost camera can estimate the distance as well, taking the rear license plate as a metric reference, including standards fix both the plate envelope and the regulated character height, so the pinhole projection converts either one into a distance. This research presents a approach and validates it. Plate localization now withstands the low-contrast and cluttered frames that previously drew the detector onto the vehicle body, or onto the character cluster of a sparsely lettered plate. Character height, plate outline, and mounting-hole span form three distance channels, merged by a consensus gated inverse-variance fusion that discards a corrupted channel before it can bias the result, with each channel separately corrected for the foreshortening of the direction along which it is measured, the correction following from the plate's recovered attitude. Finally, an innovation gate protects the temporal filter, so that a single bad detection cannot become a false warning. Evaluation is carried out in a metrology-exact testbed in which the commanded distance is the true distance, through 51 USA registries with distances of $1$ to $12$~m and viewing angles to $30^{\circ}$ under five lighting conditions. The plate is located in every frame. Mean absolute percentage error is $4.32\%$ for the outline channel, while the fusion returns $5.79\%$ mean and $2.17\%$ median, without any of its single-channel failure modes. Roll is recovered to within one degree; the two out-of-plane tilts only to within several. The algorithm can provide a solid foundation for low-cost distance estimation which can serve as an emergency backup for other sensors.
\end{abstract}


\section{Introduction}
The distance to the vehicle ahead is essential to every Longitudinal driver-assistance function. Adaptive cruise control maintains that distance, forward collision warning responds once it closes too fast, and automated emergency braking steps in when impact turns imminent~\cite{mukhtar2015vehicle,coelingh2010collision,yang2025comprehensive,su2025cooperative}. Whatever quality the range estimate has passes straight into the timing and the credibility of those interventions; a late warning is worthless, and a frequent one gets switched off by the driver. The distance is thus not an incidental detail but the footing under everything else the function does.

Production vehicles obtain the distance from an active rangefinder or from a stereo pair. Millimeter-wave radar reports range and range rate dependably through weather, but resolves the scene only coarsely. A scanning laser is dense and accurate, though it adds cost and power draw and its emitter, wheather moving or solid-state has to endure automotive services. A calibrated stereo rig recovers depth passively, but it demands a rigid baseline and a calibration that wanders under thermal cycling and vibration~\cite{yurtsever2020survey,van2018autonomous,janai2020computer}. Each one of these works well inside its own regime, and current perception stacks run several together. An inexpensive range estimate that fails differently from the active sensors around it therefore earns its place even where it is never the primary source ~\cite{kukkala2018advanced,cao2026extrinsic}.

The single ordinary camera is the cheapest sensor on the vehicle and by far the most widely fitted~\cite{liu2025electric}, yet on its own it cannot produce metric distance. That barrier is geometric, not technological. Scaling a scene size and scene depth together leaves the pinhole image untouched, so an object of unknown size at unknown range is irreducibly ambiguous~\cite{hartley2003multiple}. Learned monocular depth networks have improved quickly, but they step around that ambiguity instead of removing it by predicting depth up to an unknown global scale, or in a relative form needing a metric anchor before it can mean meters, or they take on training-set scene statistics that need not carry over to another camera or another road~\cite{eigen2014depth,fu2018deep,bhat2021adabins,ranftl2021vision,ranftl2020towards,godard2019digging}. The same limit binds monocular 3D detectors that regress a metric box per vehicle~\cite{mousavian20173d,chen2016monocular,brazil2019m3d} and pseudo-LiDAR pipelines built on a predicted depth map~\cite{wang2019pseudo}, all of them trained on large annotated driving corpora~\cite{geiger2012we,cordts2016cityscapes,caesar2020nuscenes,sun2020scalability} and all inheriting the scale sensitivity of what they predict. The classical and reliable escape is to put an object of known real size into the scene and read scale off it directly. Single-camera driver assistance has done exactly that for years with the fixed width of a lane or a vehicle~\cite{stein2003vision,dagan2004forward}, and robotics routinely drops printed referecnce of known dimension into a scene whenever metric pose is wanted~\cite{garrido2014automatic,olson2011apriltag}.

The existing methods are either confined to a marker, which may not be install on a public road, or dependent on a vehicle width that varies by model. A recent line of work escapes both by noticing that a marker of known size is already bolted to the back of every road vehicle: the license plate. USA plates are manufactured to a standardized outer envelope, and the serial characters embossed on them are set in a regulated typeface whose height each state fixes within a narrow band, so imaged size measures distance directly. Establishing the scale needs no learning at all, only the pinhole projection and the feature's physical dimension~\cite{reddy2026typography,reddy2026physics,lokeshreddy2026adas}. The plate is in effect a reference that law obliges manufacturers to attach and that owners have every reason to keep clean and visible, and since reading it also discloses the vehicle's identity, one passive sensor yields a distance, a bearing, and an identity from a single measurement. Three papers develop that idea into a working pipeline and stand as the direct ancestors of this one: the first lays out the framework and reports a coefficient of variation (CV) of $2.3\%$ in character-height measurement~\cite{reddy2026typography}, the second hardens the front end with a multi-method detector and a learned state classifier and passes a depth estimate through a Kalman filter~\cite{reddy2026physics}, and the third extends the output to bearing and vehicle identity~\cite{lokeshreddy2026adas}. Their metric backbone, which we share, is the pinhole model and its calibration~\cite{zhang2000flexible}, perspective-$n$-point pose~\cite{lepetit2009ep,terzakis2020consistently,jiang20253dgs}, robust fitting~\cite{fischler1981random}, and the standard low-level operators~\cite{canny1986computational,suzuki1985topological,duda1972use,bradski2008learning}, drawing on a mature plate-reading literature~\cite{anagnostopoulos2008license,du2012automatic,redmon2016you,shi2016end,smith2007overview,jaided2020easyocr,xu2018towards,laroca2018robust,hsu2012application} and on vehicle detection and tracking~\cite{sivaraman2013looking,ren2015faster,liu2016ssd,redmon2018yolov3,bochkovskiy2020yolov4,he2016deep,bewley2016simple,wojke2017simple}. Two of their conclusions are re-examined here. A channel, is one plate feature of known physical size, measured in the image and converted to a distance on its own, So that several channel give several independent estimates of the same distance. Their accuracy rests on a design error model and on agreement between independent channel instead of on a surveyed reference and their field set averaged barely more than a meter in range; the metrology-exact testbed of this paper supplies the external ground truth that was absent, over the full envelope. Their recommendation that character height is the preferred feature also fails to survive that wider envelope, for reasons the error analysis of Sec.~\ref{sec:system} makes plain.

However, 2 gaps separated the principle from a deployable channel, robustness and validation. Plates have to be found across a wide span of illumination, contrast, and background clutter, and a detector tuned on clean high-contrast plates breaks in two characteristic ways once the frames get harder. Put a light plate on a light vehicle and the outline is nearly edgeless, so the detector takes the largest plate-shaped region on offer instead, which is the vehicle body; the box that results is far too large and every distance drawn from it far too near. Give it a plate carrying only a few large characters and a box-tightening step meant to trim a loose detection to its densest edges can instead collapse onto the character cluster, mistaking a fraction of the plate for all of it. Either failure corrupts every downstream channel simultaneously and, once through the temporal filter raise a warning where no threat exists. As for validation, $4\%$ v.s. $5\%$ is a small difference and a decisive one when a warning threshold is being fixed, and no road test readily supplies distance ground truth at that precision across the plate designs, viewing angles, and lighting a vehicle actually meets. Neither can a road test present one plate at a controlled sequence of distances and angles so that factors can be separated. Absent that control the method's accuracy could not be mapped over the envelope it would really work in. Simulation is the natural instrument, and driving simulators are well established for perception work~\cite{dosovitskiy2017carla,goodin2022evaluating}, though what we need is narrower than a whole synthetic city: one plate whose distance is known exactly. The broader setting of camera-based assistance~\cite{bertozzi2000vision,yurtsever2020survey,van2018autonomous,janai2020computer,kukkala2018advanced} and the standards a warning function answers to frame the work throughout.

Its contributions of this paper are the following.
The ranging system itself is improved and made more robust. A character-anchored recovery and a trim guard strengthen plate localization on low-contrast and cluttered frames. The plate's three-dimensional attitude is recovered from the vanishing points of its own edges, and every ranging channel is corrected for the foreshortening of the direction it measures along. The three channels is corrected by a consensus gated, agreement aware inverse-variance rule that throws out a corrupted channel before it can bias the estimate and an innovation gate protects the temporal filter so that one mis-detection cannot become a false warning~\cite{debouk2019overview,akkus2026synoptic}

MetroSim, a metrology-exact rendering testbed, supplies the evaluation. it renders standardized plates into the image through the same pinhole geometry the methods later inverts, which makes the commanded distance an exact ground truth. It spans the full envelope of jurisdiction, distance, viewing angle and illumination in a controlled single-plate mode and a continuous driving mode alike. Re-targeting it to another camera or vehicle class amounts to editing one specification.

The rest of the paper runs as follows: Sec.~\ref{sec:system} sets out the method and its three improvements; Sec.~\ref{sec:sim} describes MetroSim, the metrology-exact testbed; Sec.~\ref{sec:results} reports and analyzes the results, and Sec.~\ref{sec:conclusion} concludes.

\section{Method}\label{sec:system}

\begin{figure}[p]
\centering\includegraphics[height=0.88\textheight,keepaspectratio]{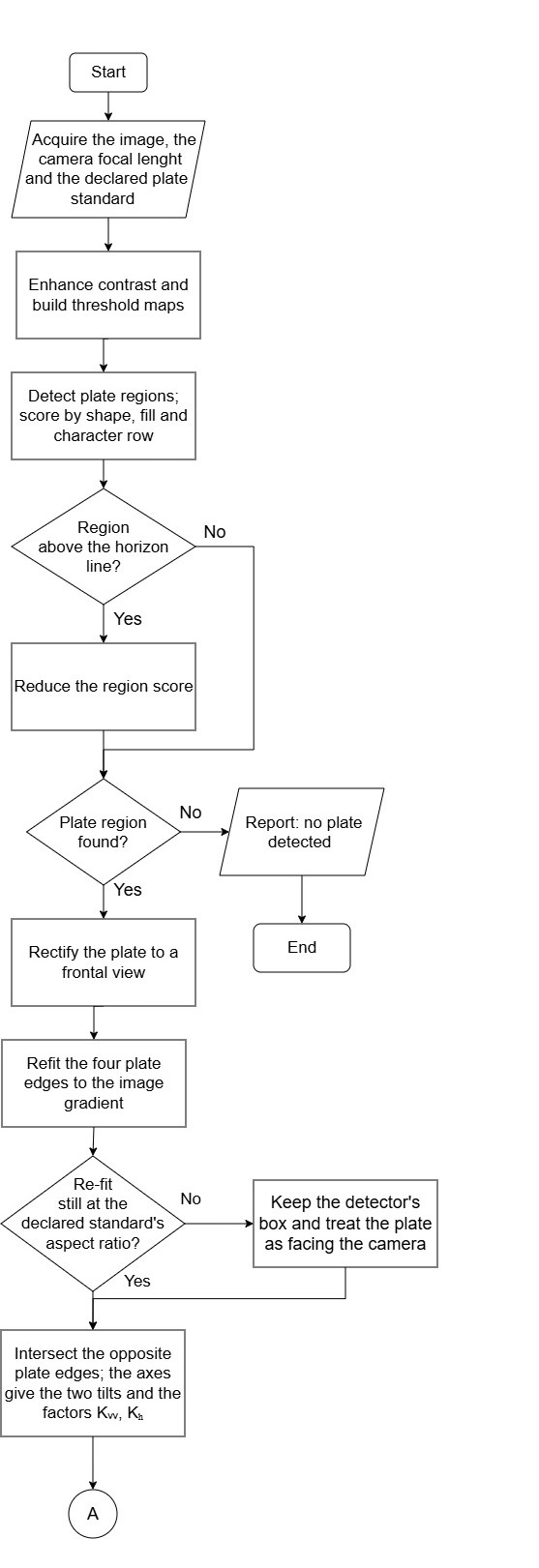}
\caption{Processing flow for one camera frame, sheet one of three: the image is acquired with a declared plate standard, the plate is located and rectified, its four edges are re-fitted to the image gradient, and intersecting the opposite edges gives the two tilts and the foreshortening factors before the sheet leaves at connector A\label{Figure:flowdet}}
\end{figure}

\begin{figure}[p]
\centering\includegraphics[height=0.88\textheight,keepaspectratio]{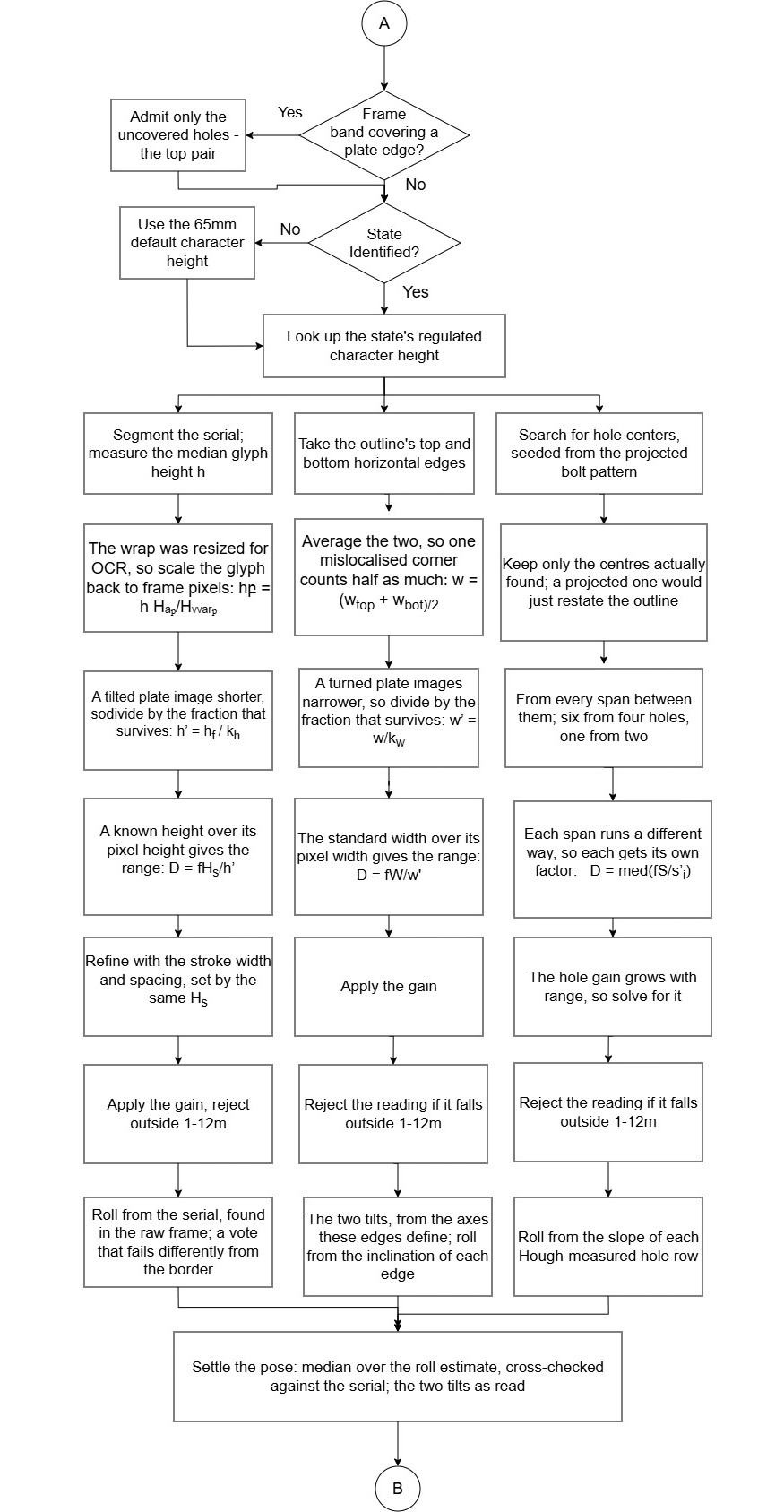}
\caption{The tree ranging channels measure in parallel and their roll estimates are settles before connector B\label{Figure:flowchan}}
\end{figure}

\begin{figure}[p]
\centering\includegraphics[height=0.88\textheight,keepaspectratio]{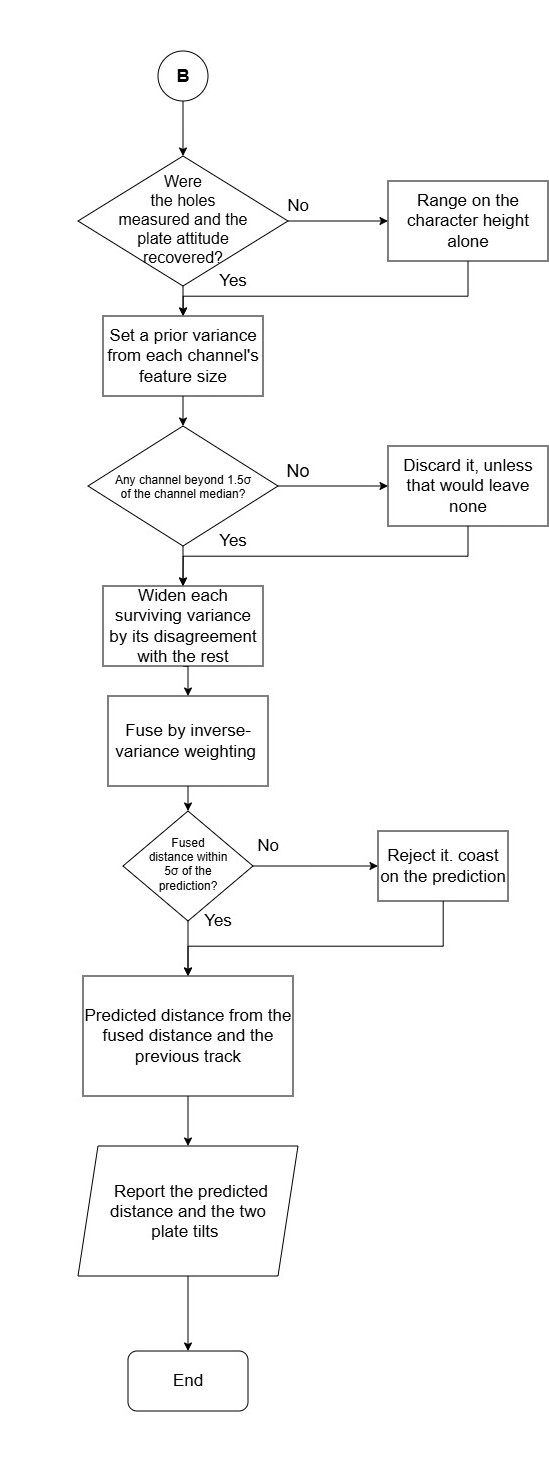}
\caption{The three distance are fused and filtered into the reported distance and the two plates tilts \label{Figure:flowfuse}}
\end{figure}

A rear license plate is treated as a target of known size, and distance follows from inverting the pinhole projection on plate features whose true dimensions the standard fixes. Figure~\ref{Figure:flowdet}, together with the two sheets after it, lays out the per-frame flow: the position of every decision, the points at which the three branches part and rejoin, and the quantity each branch measures with the standard dimension that scales it. Three sheets carry the chart instead of one, joined in order by the matched connector pairs A and B, because at the width a page allows a chart this size shrinks to roughly three-point type, and a figure the reader cannot read says nothing. Nothing is lost in the division: those both points were already joined by connectors, not flow lines under Federal Information Processing Standard (FIPS) PUB 24, which adopts American National Standards Institute (ANSI) X3.5-1970, so no flow line is severed, and each sheet now holds a stage of the method that can be discussed on its own. Sheet one locates the plate, rectifies it and recovers its attitude; sheet two identifies the jurisdiction and measures the three ranging channels in Fig.~\ref{Figure:flowchan}; sheet three fuses and filters what they return in Fig.~\ref{Figure:flowfuse}. Both warning decisions carry their thresholds outright instead of a description of them, and each has a closing-speed condition that fires independently of the time. Pose angles come from the four corner positions solved from the known plate rectangle, never from a single width, and focal length enters at acquisition because all three distances depend on it. Algorithm~\ref{alg:frame} shows the same procedure in full. Control flow is easiest to follow in figure, while the constants and the warning stage are set out in the algorithm

\begin{algorithm}[htbp]
\caption{Per-frame ranging and collision warning\label{alg:frame}}
\begin{algorithmic}[1]
\Require image frame, the running track $(\hat{D}^-, \hat{v}^-)$, focal length $f$
\Ensure fused distance $\hat{D}$, pose angles $(\theta,\psi,\phi)$, closing speed $\hat{v}$, warning level $w$
\State $\mathcal{R} \gets$ scored plate regions, with any region above the horizon row down-weighted
\If{$\mathcal{R} = \emptyset$ and a track is alive}
    \State $C \gets$ corners carried over from the previous frame
\ElsIf{$\mathcal{R} = \emptyset$}
    \State \Return no measurement this frame
\Else
    \State $C \gets$ corners of the highest-scoring region
\EndIf
\State refine $C$, then rectify the plate to a frontal view
\State $(\mathbf{R}, \mathbf{t}) \gets \textsc{SolvePnP}(C)$ \Comment{every channel below is corrected with this pose}
\State $H_s \gets$ regulated character height of the identified jurisdiction
\State $d_1 \gets f H_s / \bar{h}$ \Comment{character height}
\State $d_2 \gets f W_p / w_p$, with $W_p = 305$~mm \Comment{plate outline}
\If{the mounting holes are not detected}
    \State derive the hole positions from $C$ and $(\mathbf{R}, \mathbf{t})$
\EndIf
\State $d_3 \gets f S_b / s_b$, with $S_b = 279.6$~mm \Comment{bolt span}
\State $d_i \gets d_i / g_i(d_i)$ and $s_i \gets \sigma_i d_i$ for each channel $i$
\State $m \gets \operatorname{median}\{d_i\}$
\State $\mathcal{K} \gets \{i : |d_i - m| \le 0.40\, m\}$ \Comment{consensus gate}
\If{$\mathcal{K} = \emptyset$}
    \State $\mathcal{K} \gets \{\arg\min_i |d_i - m|\}$ \Comment{the gate never returns nothing}
\EndIf
\State $\tilde{s}_i^2 \gets s_i^2 + (d_i - \bar{d}_{\setminus i})^2$ for $i \in \mathcal{K}$
\State $z \gets \sum_{i \in \mathcal{K}} \tilde{s}_i^{-2} d_i \big/ \sum_{i \in \mathcal{K}} \tilde{s}_i^{-2}$
\If{$|z - \hat{D}^-| > \max(5\sqrt{S},\, v_{\max}\Delta t)$, with $v_{\max} = 20$~m/s}
    \State coast on the prediction, and reinitialize after six consecutive rejections
\Else
    \State $(\hat{D}, \hat{v}) \gets \textsc{KalmanUpdate}(z)$
\EndIf
\State $\tau \gets \hat{D} / |\hat{v}|$ if $\hat{v} < 0$, otherwise $\infty$
\If{$\tau < 1.5$~s or $\hat{v} < -5$~m/s}
    \State $w \gets \textsc{Danger}$
\ElsIf{$\tau < 2.4$~s or $\hat{v} < -3$~m/s}
    \State $w \gets \textsc{Caution}$
\Else
    \State $w \gets \textsc{None}$
\EndIf
\end{algorithmic}
\end{algorithm}

Localization and rectification to a frontal view of the plate come first in every frame. Three branches then run in parallel: one segments the serial characters and measures their height, one measures the width of the plate outline, and one measures the span between the mounting holes. Each branch inverts the same projection and hands back a distance of its own.

Those three distances are calibrated separately, then merged by a consensus-gated inverse-variance rule. The merged distance goes through a temporal filter that an innovation gate protects. Out of the filter come range, closing speed, time-to-collision, and a warning level.

The subsection immediately below develops what the three branches share: the projection, the behavior of its error, the per-channel calibration, and the fusion rule. The three subsections after it take the character, outline, and hole channels one at a time.

\subsection{Measurement with Calibration and Fusion}
Under the pinhole model a planar feature of true height $H$ (in meters), viewed frontally at distance $D$ along the optical axis, images to a height $h = fH/D$, with $f$ the focal length in pixels at the working image width. Distance comes from inverting that relation.
\begin{equation}\label{eqn:pinhole}
D = f\,\frac{H}{h}
\end{equation}

Every ranging channel instantiates Eq.~\eqref{eqn:pinhole}. The plate standard supplies the physical length $H$, the frame supplies the imaged length $h$, and the pair gives a distance.

Differentiating Eq.~\eqref{eqn:pinhole} shows how measurement noise reaches that distance. Because $\partial D/\partial h = -fH/h^{2} = -D/h$, a small error $\sigma_h$ in the imaged length propagates as follows.
\begin{equation}\label{eqn:relerr}
\frac{\sigma_D}{D} = \frac{\sigma_h}{h}
\end{equation}

Fractional error in distance therefore equals fractional error in imaged length. Substituting $h = fH/D$ restates this in terms of physical feature size.
\begin{equation}\label{eqn:sizelaw}
\frac{\sigma_D}{D} = \frac{\sigma_h}{f}\,\frac{D}{H}
\end{equation}

Equation~\eqref{eqn:sizelaw} is the relation the rest of the paper turns on. Hold the localization error fixed at a pixel or two and the relative distance error rises with range and falls with the size of the feature measured. A large feature is a better ruler than a small one at the same distance, and its advantage widens as the plate recedes. Roughly a fifth the width of the $305$~mm outline, the $65$~mm character should by Eq.~\eqref{eqn:sizelaw} lose accuracy about five times faster with range. Sec.~\ref{sec:results} confirms that it does.

Focal length is measured once offline, from a planar calibration target~\cite{zhang2000flexible}. It can equally be taken from the datasheet field of view as $f = (W/2)\cot(\theta_H/2)$, for image width $W$ and horizontal field of view $\theta_H$. Since $f$ enters every channel identically, moving the pipeline to another camera means editing this single number (Sec.~\ref{sec:reconfig}).

Each channel also carries a small fixed scale offset, because in practice the detector recovers a geometric quantity slightly different from the nominal dimension it is divided by. The localized outline sits just outside the printed rim and the located hole centers sit just inside the true bolt pattern. One-time calibration $d_i \leftarrow d_i/g_i$ removes each offset, the gain being the median ratio of raw reading to reference distance over the calibration set.
\begin{equation}\label{eqn:gain}
g_i = \operatorname{median}_{k}\!\left[\frac{d_i^{(k)}}{D^{(k)}}\right],\qquad d_i \leftarrow d_i/g_i
\end{equation}

Taking the median rather than the mean stops a single badly mislocalized frame from moving the gain. The gains are measured once at known distances and then held fixed, as any sensor is calibrated before use. Our sweep gives character and outline gains of $g_{\text{char}}=1.004$ and $g_{\text{outline}}=1.005$.

The hole channel is the exception, instructively so, because its offset does not stay constant with range. Bolt-hole centers are small and low in contrast, and they grow harder to locate as the plate recedes. Beyond the range at which the circle search still resolves them the channel falls back on the pose-projected bolt pattern, which carries a bias of its own. Measured across the sweep, the residual climbs almost linearly with distance, from near zero at $1$ - $3$~m to about $5\%$ at $11$ - $12$~m. No single scalar absorbs that, since any value fitted to the average over-corrects near and under-corrects far.

A first-order gain removes it.
\begin{equation}\label{eqn:holegain}
g_{\text{hole}}(D) = 0.9992 + 0.00742\,D
\end{equation}

Here $D$ is in meters. Because the gain depends on the distance being recovered, the correction is solved by a few fixed-point iterations, and the map contracts strongly enough to converge well inside the measurement noise.

We tested the form of Eq.~\eqref{eqn:holegain} before adopting its coefficients. Fitted in half the sweep and scored in the held-out half, it moved the hole channel from $7.88\%$ to $7.18\%$. Refit on the full sweep, the coefficients above gives $7.63\%$ over all $1224$ renders, with a residual bias of $-0.03\%$ and no leftover trend with range. That the held-out prediction and the full result agree to the evidence that the correction is not fitted to the frames it is judged on. The same treatment applied to the outline and to the character channel made both worse, neither residual being linear in range, so it is applied to the hole channel alone.

Five-fold cross-validation re-derives the two scalar gains to within roughly $1.5\%$ of their held-out values. They are fixed instrument offsets, not quantities tuned on the test set, and removing them before fusion is what lets the combined estimate drop below every individual channel. Because the three calibrated channels measure one distance at differing reliability then combined rather than selected. 

Every channel contributes a distance $d_i$ and a prior variance $s_i^{2}$, and that prior is not set by hand. It follows from Eq.~\eqref{eqn:sizelaw}: a fixed localization error yields a relative distance error $s_i/D \propto D/H_i$, governed by the physical size $H_i$ of the feature. So, the character height $65$~mm is several times wider than the outline $305$~mm, about $0.153D$ against $0.054D$, and is correspondingly less trusted. This is what keeps the smallest and most segmentation-sensitive feature from taking over the result as its error grows with range.

Intuition says the character channel ought to carry the most weight, the serial being the part of the plate a camera can always see where a frame may hide the holes and a low-contrast rim may blur the outline. Measurement does not bear the intuition out of this system, and it fails in availability before it fails in accuracy. The character channel was the least tested of the three across the test runs, reporting on $87.4\%$ of detected frames where the other two reported $100\%$, because it needs several glyphs segmented where the outline is the detection box itself and the hole pattern has a geometric fallback; on no frame was it the only channel available. It is also the least accurate, by about a factor of three, exactly as Eq.~\eqref{eqn:sizelaw} predicts for the smallest feature. Re-fusing the recorded channel distances with the character channel forced to dominate costs between $0.6\%$ and $2.3\%$ points of fused error depending on the sweep, so the size-derived priors stay. We also tried weighting each channel by per-frame measures of its quality instead of feature size alone and report the outcome in Sec.~\ref{sec:results}, since two of the three terms attempted made the result worse and the reasons repay more than the scheme would have.

A variance states the precision expected of a channel. It says nothing as to if that channel rests on anything observed and that second point must be settled first, because a channel with no evidence behind it deserves not a small weight but no vote. The hole channel makes that distinction concrete. When the circle search finds nothing, each hole falls back to a position projected from the pose solve, and that solve is fitted to the plate corners the same corners the outline channel measures so a span built on projected centers restates the outline instead of testing it. Admitting such a span would average one measurement twice under two names, and since the two would then agree by construction, neither the consensus test nor the inflation of Eq.~\eqref{eqn:inflate} could catch the duplication: both react to disagreement, and there would be none. The hole channel is therefore admitted only when the search actually located at least two centers, which reduces its availability from every detected frame to $88.4\%$ of them.

The duplication is measurable, but not conceivable. Correlating the relative errors of the three channels across all the test runs gives $+0.928$ between outline and hole span, $+0.651$ between character height and outline, and $+0.560$ between character height and hole span. The first is a near-duplicate, the mechanism is that the circle search is seeded from the projected bolt pattern and accepts a center only within $15$ pixels of it, so even a measured center cannot disagree much with the corners. Correlation of that size bears on the reported uncertainty, not on the estimate, since for two channels of comparable variance and correlation $\rho$ the inverse-variance rule understates the fused standard deviation by a fixed factor.
\begin{equation}\label{eqn:corr}
\frac{s_{\mathrm{true}}}{s_{\mathrm{reported}}} = \sqrt{1 + \rho}
\end{equation}

At $\rho = 0.928$ the factor is $1.39$. Little of this reaches the distance itself, two channels built on similar information returning similar answers, but a filter told that a reading is $39\%$ more certain than it is will over-trust it, and the uncertainty this system reports is meant to be usable as a filter input. We report the correlation instead of correcting for it, because the right correction is a full covariance in place of the diagonal assumed here. Sec.~\ref{sec:conclusion} records it as the largest open issue in the fusion.

The same principle settles what happens when no trustworthy attitude is recovered at all. Every channel then loses its foreshortening correction, but not equally. Outline and hole channels are measured across the plate's width, the direction an unrecovered yaw compresses, while character height is measured across its height where pitch compresses. On a road those two angles are not comparable: a lead vehicle in an adjacent lane or on a curve presents tens of degrees of yaw, whereas pitch is fixed by mounting rake and camera height and stays small. On such frames the system therefore narrows to the character channel alone. Measured on test run with the channel values held fixed and only this rule varied, the narrowing costs almost nothing on an ordinary frame and a great deal on the worst one. Median error moves from $2.26\%$ to $2.21\%$, while the 99th percentile moves from $21.1\%$ to $52.1\%$. Keeping the outline beside the character height recovers most of that tail, at $29.2\%$, the reason being that the outline is the detection box itself and so is always measured, whereas the holes are what went missing. All three settings are reported because the controlled tests cannot referee the question the rule exists for, rendering controlled contrast on a plain backing where the outline is never lost, which is in the field.

Channels reporting non-finite or out-of-band distances are discarded next. Where two or more survive, a consensus test rejects any distance more than $40\%$ from the median of the survivors, which removes any badly corrupted reading a character height measured from a merged pair of glyphs before it can move the estimate. Each surviving channel then has its variance widened in proportion to its disagreement with the rest.
\begin{equation}\label{eqn:inflate}
\tilde{s}_i^2 = s_i^2 + \bigl(d_i - \bar{d}_{\setminus i}\bigr)^2
\end{equation}

Here $\bar{d}_{\setminus i}$ is the inverse-variance mean of the other channels, so a channel slipping just under the rejection threshold is down-weighted all the same. These two variance steps serve different purpose and neither substitutes for the other. The prior states the precision expected of a channel, depending only on the size of the feature that channel measures, so it is the fusion that the $65$~mm glyph is a poorer ruler than the $305$~mm outline even on a frame where all three agree. The inflation measures the agreement between this particular reading the others, so it catches a normally precise channel corrupted on this frame. Without the prior, three agreeing channels would take equal weight though the error propagation where they should not. Without the inflation, a corrupted outline would keep its small prior variance and dominate.

The rejection threshold of $40\%$ applies to spread between channels, not to error relative to truth, and it detects a corrupted channel rather than specifying an accuracy, while fused error is $5.79\%$. The accuracy is set by the weighting that follows. Nor can it be tightened much, ordinary disagreement already being of that order. Two channels differ with a standard deviation depending on their correlation as well as on their variances.
\begin{equation}\label{eqn:diffsd}
s_{i-j} = \sqrt{s_i^2 + s_j^2 - 2\rho_{ij}\,s_i s_j}
\end{equation}

For character outline pair, at the priors above and the measured $\rho = 0.651$, this comes to $0.245D$, putting the $40\%$ threshold at $1.6$ standard deviations. The correlation term is not a refinement that may be dropped: treating the pair as independent gives $0.286D$ and $1.4$ standard deviations, so ignoring it makes the gate look looser than it is. Measured spread confirms this and shows the threshold should not be one multiple of sigma but three. About the median of the channels the quantity the gate tests the character channel scatters by $21.9\%$, the outline by $4.7\%$ and the hole span by $6.7\%$, so the same $40\%$ threshold stands at $1.8$, $8.6$ and $5.9$ standard deviations respectively. The character channel is thus the binding one, a gate having no business firing on a channel that is imprecise. Moving the threshold to $15\%$ would put it below one standard deviation there and discard the character channel on most frames for imprecision rather than error.

That reasoning is testable, and we tested it by re-fusing the recorded channel distances of the whole test cases at a range of thresholds. Over the entire span from $5\%$ to no gate at all the fused error moves by $0.26\%$  points, and removing the gate altogether costs $0.14$. The gate is therefore not what delivers the accuracy, which answers the apparent contradiction: a loose rejection threshold and a small final error coexist because the work is done by the inverse-variance weighting and not by the threshold. A channel surviving at a $39\%$ deviation has its variance inflated by the square of that deviation and contributes almost nothing. On this sweep $40\%$ happens to be the best of those tested the fused error being $4.24\%$ at that threshold, $4.26\%$ at $30\%$ and $4.25\%$ at $50\%$. We attach little weight to that, the differences sitting inside the noise of a single sweep and this sweep containing no corrupted channels for the gate to act on, so tuning the threshold here would fit the clean case the gate does not exist to handle. Survivors are then combined by inverse-variance weighting.
\begin{equation}\label{eqn:fuse}
\hat{D} = \frac{\sum_i w_i\,d_i}{\sum_i w_i}, \qquad w_i = \frac{1}{\tilde{s}_i^2}
\end{equation}

A two-state filter then tracks the fused distance, its state $\mathbf{x} = [D,\; \dot{D}]^{\mathsf{T}}$ holding range and range rate. To allow for a braking or accelerating target, the process noise is sized from a physical acceleration disturbance of standard deviation $\sigma_a$ rather than from an isotropic constant.
\begin{equation}\label{eqn:Q}
\mathbf{F} = \begin{bmatrix} 1 & \Delta t \\ 0 & 1 \end{bmatrix},\qquad
\mathbf{Q} = \sigma_a^2 \begin{bmatrix}\tfrac{1}{4}\Delta t^4 & \tfrac{1}{2}\Delta t^3 \\[2pt]\tfrac{1}{2}\Delta t^3 & \Delta t^2 \end{bmatrix}
\end{equation}

Here $\sigma_a = 4$~m/s$^2$, a value spanning ordinary urban braking. Measurement noise is not constant but grows with range in step with the propagation of Eq.~\eqref{eqn:pinhole}, $R(D) = (c_v D)^2$ with $c_v$ the relative error of the fused input. A measurement is admitted only if its innovation $\nu = z - \mathbf{H}\mathbf{x}^-$ lands inside a gate set by the wider of a statistical and a physical bound.
\begin{equation}\label{eqn:gate2}
|\nu| \;\le\; \max\!\left(N_\sigma\sqrt{S},\;v_{\max}\,\Delta t\right),\qquad S = \mathbf{H}\mathbf{P}^-\mathbf{H}^{\mathsf{T}} + R
\end{equation}

The second term expresses that no vehicle can change range by more than $v_{\max}\Delta t$ between frames. Time-to-collision and warning level follow from the filtered state.
\begin{equation}\label{eqn:ttc}
\begin{gathered}
\tau = \frac{D}{|\dot{D}|}\quad (\dot{D}<0),\\[2pt]
W =
\begin{cases}
\text{danger}: & \tau < 1.5~\text{s} \lor \dot{D} < -5~\text{m/s} \\
\text{caution}: & \tau < 2.4~\text{s} \lor \dot{D} < -3~\text{m/s} \\
\text{none}: & \text{otherwise}
\end{cases}
\end{gathered}
\end{equation}

The fused estimate reports the uncertainty $\hat{\sigma} = (\sum_i w_i)^{-1/2}$. That reported uncertainty is itself a claim, and one the downstream filter acts on the priors are set to withstand testing instead of being stated: each channel's prior follows the Gaussian relation between mean absolute deviation and standard deviation, $E|e| = \sigma\sqrt{2/\pi}$, making an honest prior the channel's measured relative deviation divided by $0.798$. Sec.~\ref{sec:results} verifies the result directly. With the channels calibrated and weighted by their propagated precision, the fusion is both more accurate than any single channel and self-checking. Three features failing in different ways making their mutual disagreement a usable signal; the consensus gate is what stops two channels that fail together from out-voting a lone correct one through their small prior variances.

Finally the fused distance is smoothed across frames by a constant-velocity Kalman filter~\cite{kalman1960new,welch1995introduction} whose measurement noise comes from $\hat{\sigma}$, so a low-confidence frame moves the state less. The filter maintains distance and closing speed and derives time-to-collision from their ratio. A box that momentarily jumps to the vehicle body reads a distance far from truth, and an unguarded filter would chase it toward a physically impossible value and might raise a spurious warning. We add an innovative gate writing $\hat{z}$ for the predicted measurement and $S$ for its innovation covariance, the acceptance test on a measurement $z$ takes the following form.
\begin{equation}\label{eqn:gate}
|z - \hat{z}| \le \max\!\bigl(N_\sigma\sqrt{S},\; v_{\max}\,\Delta t\bigr)
\end{equation}
where $N_\sigma$ is a confidence multiple and $v_{\max}\Delta t$ the largest physically plausible inter-frame range change. An out-of-gate measurement is rejected and the filter coasts on its prediction. Re-initializing onto the measurement only when the disagreement persists across several frames, which signals a genuine change of target rather than a transient error. A forward collision warning is raised once time-to-collision drops below the thresholds. A relative-depth network~\cite{ranftl2020towards} runs at reduced rate and anchored to the fused distance purely as a consistency check; it is never allowed to set the scale, which would bring back the ambiguity the plate was chosen to avoid.

\subsection{Three-Dimensional Pose and Foreshortening}\label{sec:pose}

\begin{figure}[htbp]
\centering\includegraphics[width=\linewidth]{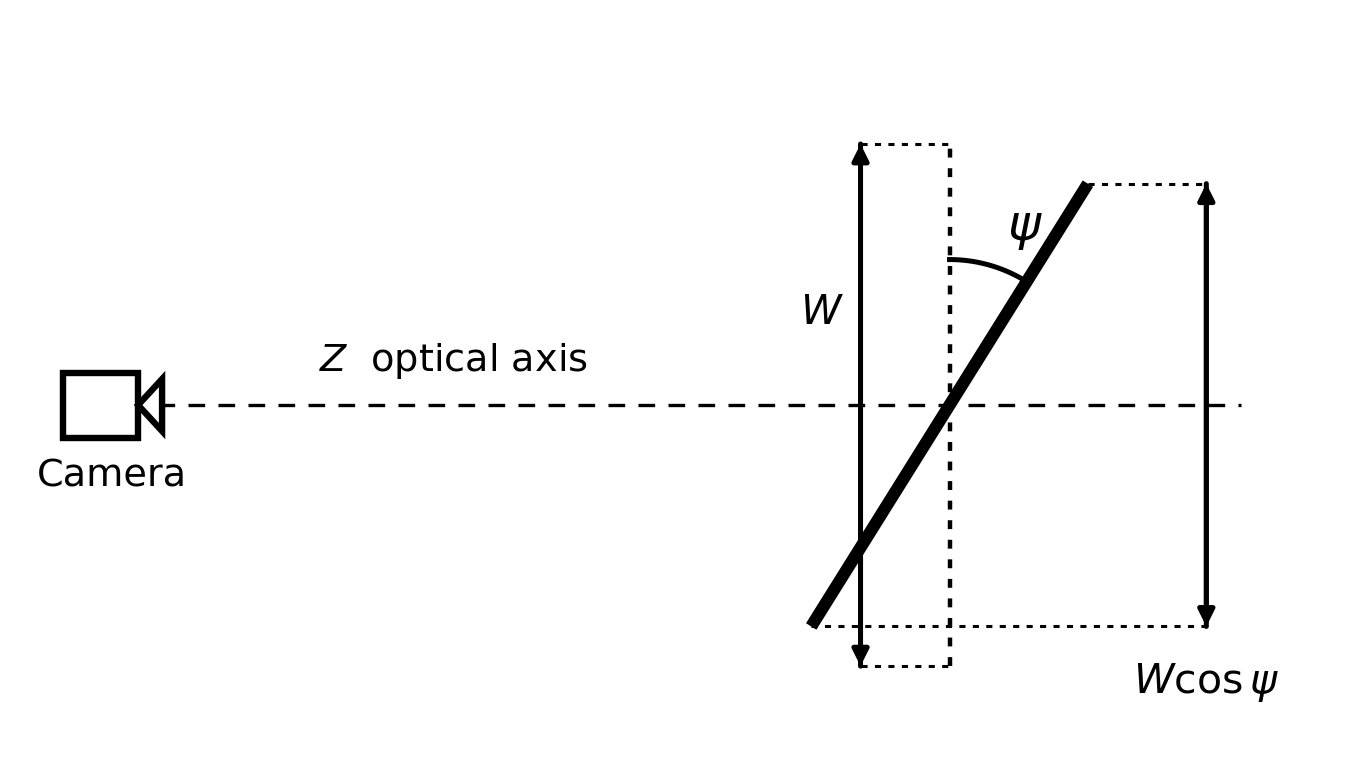}
\caption{An oblique plate images shorter, by an amount that depends on the direction measured\label{Figure:short}}
\end{figure}

Equation~\eqref{eqn:pinhole} assumes the measured feature lies in a plane facing the camera, which a plate on a vehicle ahead rarely does. The plate turns about the vertical axis when the lead vehicle occupies an adjacent lane or a curve, and about the horizontal axis through mounting rake and through the camera sitting above or below plate height. Either rotation shortens the imaged feature without moving the plate, so the naive inversion reports too large a distance. Figure~\ref{Figure:short} shows the mechanism in plan view. Write $\psi$ for horizontal tilt, $\theta$ for vertical, and $R$ for the plate's rotation in camera coordinates, and the plate's own width and height axes map to the unit vectors $\mathbf{u} = R\hat{\mathbf{x}}$ and $\mathbf{v} = R\hat{\mathbf{y}}$. Projection throws away whatever component of a direction lies along the optical axis, so only the perpendicular part survives into the image. That surviving fraction is the axis's foreshortening factor.
\begin{equation}\label{eqn:kfactor}
k_w = \sqrt{u_x^2 + u_y^2},\qquad k_h = \sqrt{v_x^2 + v_y^2}
\end{equation}

Each ranging channel divides by its own factor, since each measures along a different direction on the plate. A segment lying along $(d_x, d_y)$ in the plate's own axes has camera-frame direction $\mathbf{d} = d_x\mathbf{u} + d_y\mathbf{v}$ and factor $k = \lVert(d_x^{\mathrm{cam}}, d_y^{\mathrm{cam}})\rVert / \lVert\mathbf{d}\rVert$. The outline channel measures the plate's width and so takes $k_w$; the character channel measures glyph height and takes $k_h$; horizontal hole spans take $k_w$, vertical spans $k_h$, and the two diagonals take neither, a diagonal being foreshortened by a mixture of both. The corrected form of Eq.~\eqref{eqn:pinhole} divides the imaged length by the factor belonging to the direction it was measured along.
\begin{equation}\label{eqn:pinholecorr}
D = f\,\frac{H\,k}{h}
\end{equation}

Table~\ref{tab:kfactors} lists every length the system measures, the direction it lies along the plate, and the factor that follows. Eight measurements come from our four features and they resolve into three distinct factors rather than one. Writing a single correction for all of them is natural enough if the plate is treated as a flat object at a distance. That correction would be right for the outline, wrong by $\cos\theta/\cos\psi$ for the character height and wrong by an intermediate amount for each diagonal.

\begin{table}[htbp]
\caption{Every measured length, plate-frame directions and foreshortening factors from Eq.~\eqref{eqn:kfactor} \label{tab:kfactors}}
\centering\small
\begin{tabular}{@{}llll@{}}
\hline
measurement & $H$ (mm) & $(d_x,d_y)$ & $k$ \\
\hline
character height   & $H_s\approx65$ & $(0,1)$ & $k_h$ \\
plate outline width & $305.0$ & $(1,0)$ & $k_w$ \\
holes, top row      & $279.6$ & $(1,0)$ & $k_w$ \\
holes, bottom row   & $279.6$ & $(1,0)$ & $k_w$ \\
holes, left column  & $126.6$ & $(0,1)$ & $k_h$ \\
holes, right column & $126.6$ & $(0,1)$ & $k_h$ \\
holes, diagonal     & $306.9$ & $(+279.6,\,126.6)$ & mixed \\
holes, anti-diagonal & $306.9$ & $(-279.6,\,126.6)$ & mixed \\
\hline
\end{tabular}
\end{table}

Measured lengths, plate-frame directions, and foreshortening factors from Eq.~\eqref{eqn:kfactor}. Two properties of Eq.~\eqref{eqn:kfactor} bound what the correction can and cannot do. The factor is written on the full rotation $R$, not on the two tilts, as it must be: on a tilted plate the directions in its plane foreshorten by differing amounts, so rolling the plate turns its width axis into a direction with a different factor. Hold the normal fixed at $(0.321,-0.342,0.883)$ and vary roll alone from $0^{\circ}$ to $45^{\circ}$, and $k_w$ moves from $0.940$ to $1.000$ while $k_h$ moves from $0.947$ to $0.883$. Two angles fix which way the plate faces; foreshortening a named direction on it takes three. The factor is also bounded below by geometry and not by a clamp, since $k\to0$ only as the plate edge-on turns, where no channel can measure it at all and the decomposition of $R$ is itself singular, so the limit that breaks the correction lies outside the envelope that the method operates in.

Table~\ref{tab:cases} summarizes the conditions this correction can meet and the system's response in each. It appears as a table rather than described in the text because a correction right in the ordinary case and unexamined elsewhere is not a contribution, and because the two failure directions are asymmetric: an uncorrected channel is wrong by a bounded amount that Eq.~\eqref{eqn:kfactor} predicts, whereas a channel corrected by a wrong angle is wrong by an unbounded one. Every entry that withholds the correction withholds it for that reason.

\begin{table}[htbp]
\caption{Every condition the pose recovery and foreshortening correction can meet the system's response in each, and its measured cost on the controlled sweep\label{tab:cases}}
\centering\small
\begin{tabular}{@{}p{0.27\textwidth}p{0.40\textwidth}p{0.27\textwidth}@{}}
\hline
Condition & System's function & Measured cost \\
\hline
Plate frontal, $\psi=\theta=0$ & Vanishing points lie at infinity; the homogeneous cross product of Eq.~\eqref{eqn:vp} stays finite and returns $k=1$ & None; $\mathrm{d}k/\mathrm{d}\psi=0$ here, so the correction is best conditioned exactly where the angle is worst \\
Plate oblique, quadrilateral trusted & Each length divided by the factor of its own direction & Outline bias $+3.7\%$ at $30^{\circ}$ \\
Quadrilateral fails the aspect test of Eq.~\eqref{eqn:aspect} & Correction withheld; plate treated as frontal; variance widened & Outline bias $+8.9\%$ at $30^{\circ}$; accepted on $59.3\%$ of frames \\
Detector returns a near-rectangle & Edges re-fitted to the image gradient before the vanishing points are formed & Without it, $-0.23^{\circ}$ recovered for a true $30^{\circ}$ \\
Plate rolled as well as tilted & $k$ computed on the full rotation $R$, not on $\psi,\theta$ alone & $6\%$ if roll were dropped \\
Plate turned edge-on, $\cos\psi\to0$ & Out of operating envelope; no channel resolves the plate & Not applicable \\
Frame band covers a plate edge & Covered holes excluded from the span set; outline variance widened, not corrected, since the lip depth is unobservable & Outline median $2.24\%\to4.61\%$ \\
Jurisdiction not identified & Standard $65$~mm character height used & $0.41\%$ points of fused error \\
Plate of non-standard physical size & Not inferred from geometry; declared per vehicle class & $1.71\times$ too far if undeclared \\
Fewer than three glyphs segmentable & Character channel absent; the other two continue & Character channel available on $87.4\%$ of frames \\
Mounting holes not resolved & Positions interpolated from the plate corners and flagged as not measured, so they cannot vote on roll & Roll falls back to the outline estimate alone \\
One channel corrupted & Consensus gate drops it at $40\%$ from the median; agreement inflation collapses the weight of any that survives & Gate fires on $5.1\%$ of readings \\
Mis-detection reaches the filter & Innovation gate rejects beyond $5\sigma$; filter coasts; re-initializes after six consecutive rejections & Contains a worst single-frame range error of $370\%$ \\
\hline
\end{tabular}
\end{table}

Recovering $\psi$ and $\theta$ needs no knowledge of the distance, and that is what makes the correction possible at all. Perspective-$n$-point solving has to separate rotation from translation simultaneously, and for a target as small and as nearly frontal as a plate the two are coupled, so an error in recovered depth surfaces as an error in recovered tilt. Orientation on its own is better posed. In three dimensions the plate's two horizontal edges are parallel and meet at the vanishing point of its width axis, its two vertical edges meet at its height axis, and a vanishing point fixes a direction without reference to depth or scale. Forming each as the intersection of two image lines in homogeneous coordinates keeps the construction finite even for a frontal plate, whose vanishing points lie at infinity.
\begin{equation}\label{eqn:vp}
\mathbf{p}_w = (\mathbf{x}_{\mathrm{tl}}\times\mathbf{x}_{\mathrm{tr}})\times(\mathbf{x}_{\mathrm{bl}}\times\mathbf{x}_{\mathrm{br}}),\qquad
\mathbf{p}_h = (\mathbf{x}_{\mathrm{tl}}\times\mathbf{x}_{\mathrm{bl}})\times(\mathbf{x}_{\mathrm{tr}}\times\mathbf{x}_{\mathrm{br}})
\end{equation}

Back-projecting through the intrinsics gives the axes as $\mathbf{u} \propto K^{-1}\mathbf{p}_w$ and $\mathbf{v} \propto K^{-1}\mathbf{p}_h$, and a rectangular plate requires $\mathbf{u}\cdot\mathbf{v} = 0$, so the residual non-orthogonality of the measured pair is removed by rotating each axis half way toward the other about their common normal. The plate normal $\mathbf{n} = \mathbf{u}\times\mathbf{v}$ completes the rotation $R = [\,\mathbf{u}\;\mathbf{v}\;\mathbf{n}\,]$, from which $\psi$ and $\theta$ follow by the decomposition $R = R_y(\psi)R_x(\theta)R_z(\phi)$. We state the composition order because it has to match the order in which the testbed of Sec.~\ref{sec:sim} applies the angles it commands, or a reported angle and a commanded angle would not be the same quantity and the test would score a difference of convention as an error. In-plane roll does not come from this construction. Roll is the one angle needing no perspective, being the inclination of the plate's horizontal edges within the image, and reading it straight off those edges measures it about four times more accurately than routing it through Eq.~\eqref{eqn:vp}, where it inherits the conditioning of the two harder angles.

Roll is also the one angle with more than one independent estimator, so it gets the same robust aggregation the hole spans get rather than a single reading. The plate's outline supplies one estimate through its edge inclinations; each row of mounting holes supplies another, the holes forming a rigid rectangle on the plate so that the image angle of a row is the plate's roll. The reported roll is the median of whichever are available.
\begin{equation}\label{eqn:rollmed}
\phi = \operatorname{med}\bigl\{\phi_{\mathrm{outline}},\;\phi_{\mathrm{row,top}},\;\phi_{\mathrm{row,bot}}\bigr\}
\end{equation}

That means three estimates on an unframed plate with resolved holes, two when a frame covers one hole row, and one when the holes are not resolved at all. Only a row whose two centers the circle search actually located may vote. A row of positions projected from the pose adds nothing independent it was placed by the attitude it would be voting on and admitting it would look like corroboration while contributing none, and would outvote the one independent estimate whenever both rows were projections.

The two tilts deliberately receive no such treatment, and the asymmetry is not an oversight. Their only alternative estimator is the perspective-$n$-point rotation, which measures $5.25^{\circ}$ and $5.43^{\circ}$ mean absolute error relative to a construction exact on true corners. Since a median of two values is their mean, combining a good estimate with a known-poor one would reduce the results. Robust aggregation needs estimators of comparable quality, and for pitch and yaw we have no second one.

Given the plate's true projected corners this construction is exact. On synthetic projections it returns all three angles to within $10^{-11}$ degrees and cuts the outline channel's error at $45^{\circ}$ of horizontal tilt from $41.4\%$ to $0.01\%$. Supplied with the corners a blob-based detector actually produces it is not exact, and the reason lies in that detector, not in the geometry. A box fitted to a morphologically closed blob is nearly a rectangle, and a rectangle has both pairs of edges parallel, so its vanishing points lie at infinity and it asserts by construction that the plate is frontal. The perspective information the pose needs has been thrown away before the pose is computed. Compared with ground truth, the box sits $5$ to $22$~pixels from the true corners and returns $-0.23^{\circ}$ of horizontal tilt for a plate genuinely turned by $30^{\circ}$.

Recovering it means re-fitting the quadrilateral to the image gradient before use. A plate edge is the boundary between plate face and body panel behind it, and it images as a straight line under any pose. Each of the four edges is therefore re-estimated by sampling across it, locating the intensity step to sub-pixel precision, and fitting a line with outlier rejection, after which intersecting consecutive lines gives corners lying on the true projected quadrilateral. Two details make this work instead of fail. The step taken is the outermost one above a fraction of the profile's strongest, not the strongest itself, because the characters inside the plate are higher in contrast than the plate's border on dull sheet metal, so seeking the strongest step reliably shrinks the box onto the serial. The result is then accepted only if it is consistent with a plate having been seen from somewhere, a test the standard supplies that holds under any pose and costs nothing further to evaluate.
\begin{equation}\label{eqn:aspect}
\frac{w/k_w}{h/k_h} = \frac{W_{\mathrm{plate}}}{H_{\mathrm{plate}}} = \frac{305}{152}
\end{equation}

The ratio in Eq.~\eqref{eqn:aspect} is a property of the declared standard, not of the jurisdiction, and the two claims have to be separated because only one of them holds. Every USA jurisdiction issues its passenger plate on the same $305\times152$~mm blank, standardized in 1957, so the ratio does not vary from state to state: all $57$ synthetic renders in our library measure $2.007$ exactly. The $75$ real photographs in the same library scatter from $1.880$ to $2.163$, yet their median of $1.990$ falls within $0.9\%$ of nominal, which places that spread in where each photograph was cropped, not in what the plates are. Either way the tolerance has to absorb it, so the test is a plausibility band and not an equality. The one thing the ratio will not tolerate is the change of vehicle class. A motorcycle plate is $178\times102$~mm, an aspect of $1.745$, and Sec.~\ref{sec:sim} reports why the band cannot exclude it without excluding more passenger plates than motorcycles.

A refinement violating Eq.~\eqref{eqn:aspect} beyond a tolerance is discarded and the correction is not applied at all. This last point matters more than it looks. Uncorrected, the outline channel is biased by a known and bounded amount; corrected by a wrong angle, it is biased by an unbounded one, and an untrustworthy quadrilateral was measured to yield $k_w = 0.56$ where the truth is $1.0$, which would have injected a $79\%$ error into a channel otherwise accurate to a few percent. Declining to correct is the safe failure, and the frames on which the system declines are recorded and their variance widened instead of being passed off as frontal.

\subsection{The Character Channel and Serial Measurement}
The plate has to be found before a distance can be read from its characters, and making that step reliable is the first of the three improvements.

The detector builds several thresholded views of the frame at once: adaptive Gaussian thresholding, a global Otsu threshold, an intensity-boosted view for dark plates, and a color-masked view isolating high-chroma serial text, each taken after contrast-limited adaptive histogram equalization. Border-following contour analysis then extracts candidate quadrilaterals from every view~\cite{suzuki1985topological,canny1986computational,bradski2008learning}. Each candidate takes a score combining these cues, damped when it sits above the horizon row $y_h$.
\begin{equation}\label{eqn:score}
S = S_{\mathrm{ar}}\,S_{\mathrm{fill}}\,S_{\mathrm{char}}\cdot\begin{cases} \lambda, & \bar{y} < y_h,\\ 1, & \bar{y} \ge y_h,\end{cases}\qquad \lambda = 0.35
\end{equation}

Here $\bar{y}$ is the candidate's centroid row. The penalty encodes a simple physical fact: for any camera mounted at or above plate height, a plate on a vehicle ahead projects below the horizon, while building faces and signboards sit above it, and being rectangular and often close to the plate's $2\text{:}1$ aspect they are the decoys that matter. The penalty is a weight and not a rejection, so a high-mounted plate on a heavy vehicle seen from a low bumper camera can still win when it is the only candidate on offer. The best candidate is refined to sub-pixel corners and rectified by the homography mapping its corners onto a canonical rectangle~\cite{hartley2003multiple}.

Two failure modes on hard frames drove this improvement. The first shows up when plate-to-background contrast is low, since no single threshold then isolates the outline. We therefore also build plate candidates from the character row, which stays high in contrast even where the rim does not: the glyph row is measured and the surrounding plate inferred from known typographic proportions. These candidates compete in the same scoring, so a character-anchored candidate wins when the outline itself offers no usable edge, and the detector recovers the plate instead of the vehicle body.

\begin{figure}[htbp]
\centering\includegraphics[width=0.82\linewidth]{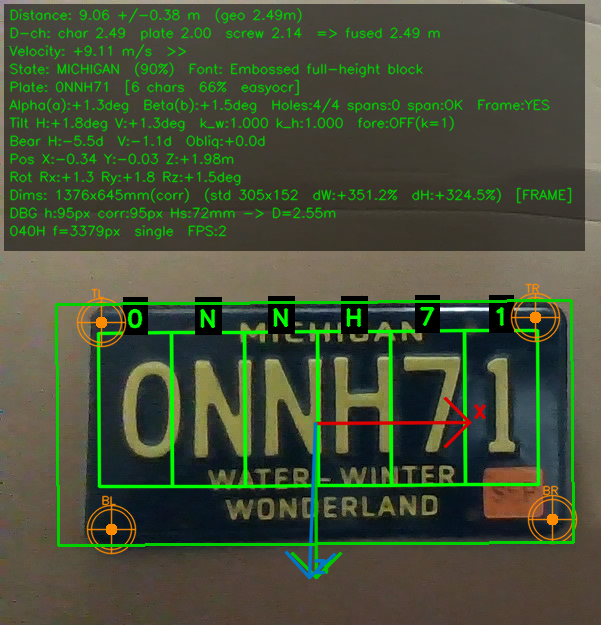}
\caption{The three ranging channels measured on one detected plate\label{Figure:detection}}
\end{figure}

The second failure mode lives in the box-tightening step that trims a loose detection to its densest edges. On a sparse plate that trim can collapse onto the character cluster and discard the rim. We disable it once a box already reads as a full plate, which preserves the outline. Figure~\ref{Figure:detection} illustrates the result on an oblique plate: the recovered outline in green, the located mounting holes in amber, and the rectified serial with each glyph boxed in blue, the three channels agreeing to within a few hundredths of a meter of the true four meters.

With the plate rectified, the character channel measures the serial. USA jurisdictions draw the serial in the Federal Highway Administration (FHWA) highway alphabet at a regulated height $H_s$ varying only from about $63$ to $72$~mm across states. The serial is segmented into per-character boxes, the median glyph height measured, de-foreshortened by the recovered pose (Sec.~\ref{sec:frame}), and put into Eq.~\eqref{eqn:pinhole} with $H = H_s$.

The same segmentation yields two further typographic quantities at no extra cost. Stroke width sits near one-sixth of character height, and inter-character spacing near one-fifth. Both are combined with the height by inverse-variance weighting inside the channel, and the height dominates, as Eq.~\eqref{eqn:sizelaw} says it should, being the largest of the three.

The issuing jurisdiction sets $H_s$ and is identified by a weighted vote over four readers that fail independently: the state name or slogan read as text, a learned appearance model, a convolutional classifier over the whole plate, and the dominant background color, with the serial format narrowing the candidate set before the appearance readers vote within it. That redundancy is not decoration, since two ordinary conditions remove the strongest reader outright. A frame covering the plate's top edge hides the header where the state name sits, which carries the highest vote weight; and a specialty, collegiate or vanity design may carry no legible state text at all, of which the database records several hundred variants across the 51 jurisdictions. The vote completes on the remaining three readers in both cases, and when none of them settles it, the standard $65$~mm height is used.

That degradation is tolerable because of the fusion weighting, not because the identification is reliable. Measuring the smallest feature, the character channel takes a prior variance far wider than the other two, which leaves it $4.26\%$ of the fused weight. Re-fusing our sweep with the largest height error the database admits Michigan's $72$~mm read as the $65$~mm default, a $10.8\%$ scale error on that channel moves the fused result from $5.88\%$ to $6.29\%$, a penalty of $0.41\%$ points. Jurisdiction matters for the identity the system reports and barely reaches the range.

A plate of different physical size, as opposed to different design, is the exception. USA motorcycle plates are $178\times102$~mm, and every channel divides by the passenger constants, so such a plate reads about $305/178 = 1.71$ times too far under-reporting range, which is the dangerous direction. We found no reliable discriminator from plate geometry alone, the outline-to-hole disagreement on a scaled plate falling inside the normal spread for passenger plates. The plate standard therefore has to be declared for the target class or taken from upstream vehicle classification; we do not infer it from the image.

Plates on the road vary a great deal, and several cases are handled explicitly.

Touching or merged glyphs are common on tightly kerned serials. A vertical-projection valley search splits them, so a joined pair is not measured as one oversized character. Punctuation, embossed state silhouettes and stacked verification stickers pose the opposite problem: aspect ratio and position reject them as non-glyph components, and they never enter the height estimate.

Vanity and graphic-background plates carry artwork that can rival the serial in contrast, and the character-row scoring anchoring localization handles them without special treatment. Dark-on-dark and bezelled designs are the hardest case and the one that most degrades the geometric channels; the intensity-boosted and color-masked views are how they are reached.

Serial length imposes no assumption. Being a per-glyph statistic, the height does not depend on how many characters there are, so a short vanity string and a full seven-character serial are treated alike. Specialty and motorcycle plates use smaller formats and are ranged by whichever channels their reduced feature set still supports.

The character channel measures the smallest of the three features, which is the richest in information and the least accurate at range, exactly as Eq.~\eqref{eqn:sizelaw} predicts. Its role is a close-range refinement and an independent cross-check, not the primary ruler.

\subsection{The Outline Channel and Plate-Frame Geometry}\label{sec:frame}
The most accurate channel reads distance from the plate's outer frame instead of from its text. North American passenger plates are standardized at $305\times152$~mm, so the imaged width of the rectified outline converts straight through Eq.~\eqref{eqn:pinhole} with $H = 305$~mm.

Behind that measurement sits the plate's full six-degree-of-freedom pose, on which the other two channels also depend for their pose correction. The pose is recovered by solving the perspective-$n$-point problem on the four physical corners~\cite{lepetit2009ep,terzakis2020consistently}, EPnP supplying an accurate linear-time initialization and SQPnP a globally optimal refinement. Random sample consensus (RANSAC)~\cite{fischler1981random} rejects a corner that has been mislocated.

Whenever the plate is turned its imaged outline width is foreshortened, so what enters Eq.~\eqref{eqn:pinhole} is the frontal-equivalent width $w_{\mathrm{meas}}/k_w$ of Sec.~\ref{sec:pose} rather than the raw imaged length. Two properties of that correction are easy to get backwards. The first is that roll cannot be discarded, tempting though discarding it is. On a frontal plate every in-plane direction runs parallel to the image plane, so all of them survive projection intact and roll does not matter. On a tilted plate it does, the plate's plane then being oblique to the image plane so that directions within it foreshorten by differing amounts, which is why Eq.~\eqref{eqn:kfactor} is written on the full rotation, not on the tilts, at the cost quantified in Sec.~\ref{sec:pose}. The second property is that the sign of the correction is fixed. An uncorrected reading is too far and not too near, a turned plate imaging narrower and a narrower plate being read as a more distant one. Measured on our sweep before the correction, a plate turned by $30^{\circ}$ read $13.2\%$ farther than it is.

The pose comes from minimizing the reprojection residual of the four known plate corners $\mathbf{X}_i$ against their detected image points $\mathbf{u}_i$.
\begin{equation}\label{eqn:pnp}
(\mathbf{R}^\star,\mathbf{t}^\star) = \arg\min_{\mathbf{R},\mathbf{t}} \sum_{i=1}^{4} \bigl\lVert \mathbf{u}_i - \pi\bigl(\mathbf{R}\mathbf{X}_i + \mathbf{t}\bigr)\bigr\rVert^2
\end{equation}
where $\pi(\cdot)$ is the pinhole projection. Four coplanar corners are the minimum this problem admits, so a mislocated corner matters and RANSAC is used.

The rotation $\mathbf{R}^\star$ holds the three angles, read from its entries in closed form. Writing $R_{jk}$ for the entry in row $j$ and column $k$, the ZYX decomposition gives the following.
\begin{equation}\label{eqn:euler}
\begin{gathered}
\psi = \arcsin\bigl(-R_{31}\bigr) \qquad
\theta = \operatorname{atan2}\bigl(R_{32}, R_{33}\bigr) \\[2pt]
\phi = \operatorname{atan2}\bigl(R_{21}, R_{11}\bigr)
\end{gathered}
\end{equation}

This decomposition is singular only at $\cos\psi = 0$, the plate turned fully edge-on, an attitude at which no channel can measure it anyway. The pose solve is retained for the translation and for steering the hole search, but it is not what supplies the reported angles: the two tilts come from the vanishing-point construction of Sec.~\ref{sec:pose} and the roll from the plate's own edge inclinations, each taken from the estimator well posed for it. Nor is any angle estimated more than once. An earlier arrangement computed plate angle in five places by four methods, and the ranging channels used none of them, so the attitude displayed and the attitude the distance was corrected by were unrelated quantities. Consolidating them means a disagreement between the two independent estimates that remain, the outline's roll and the roll implied by the mounting-hole rows, is acted on instead of merely reported: a failed cross-check widens the variance of every pose-dependent channel on that frame.

\begin{figure}[htbp]
\centering\includegraphics[width=\linewidth]{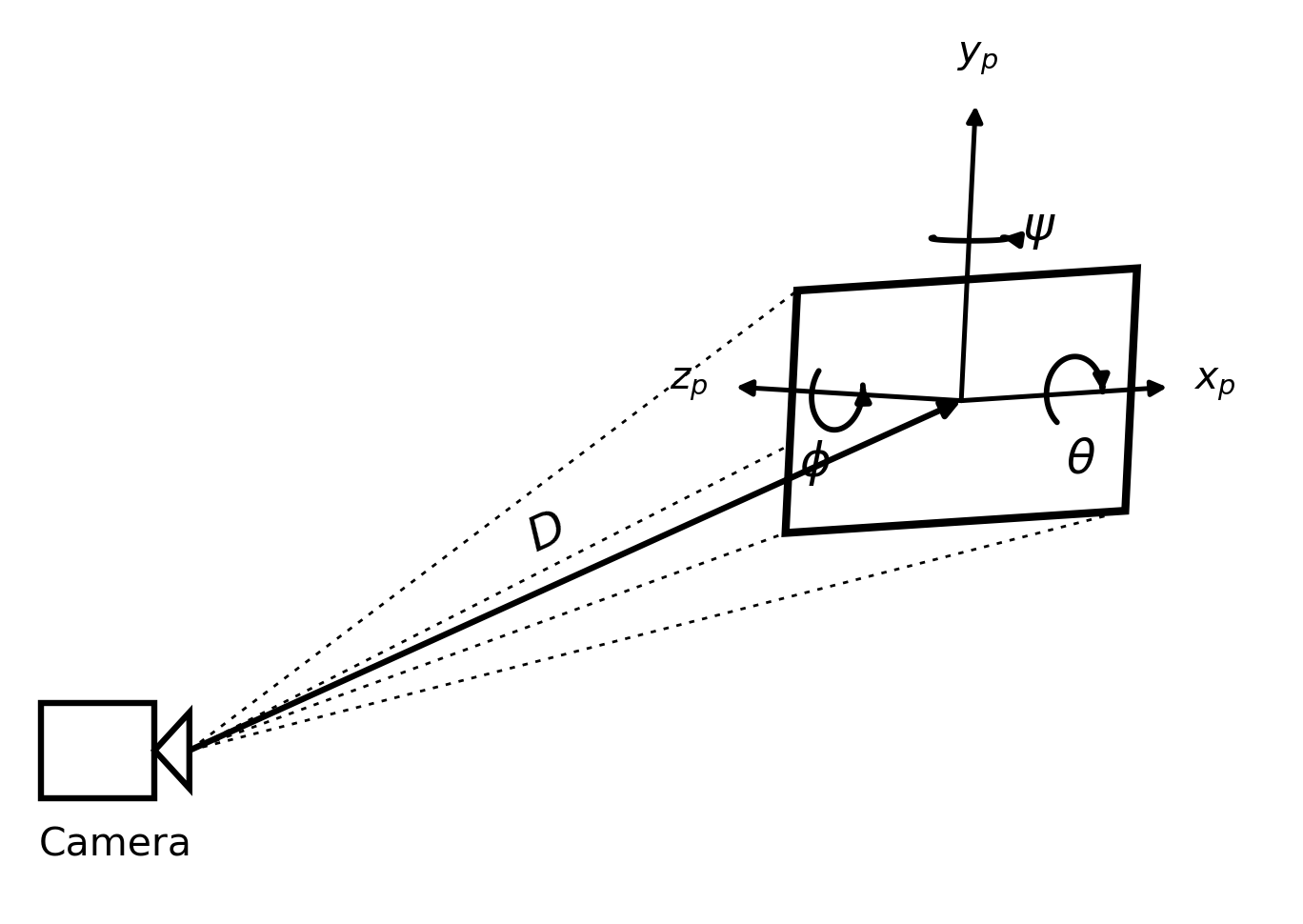}
\caption{The measurement geometry: the range $D$ to the plate, and the three rotations named on the plate's own axes\label{Figure:geom}}
\end{figure}

\begin{figure}[htbp]
\centering\includegraphics[width=\linewidth]{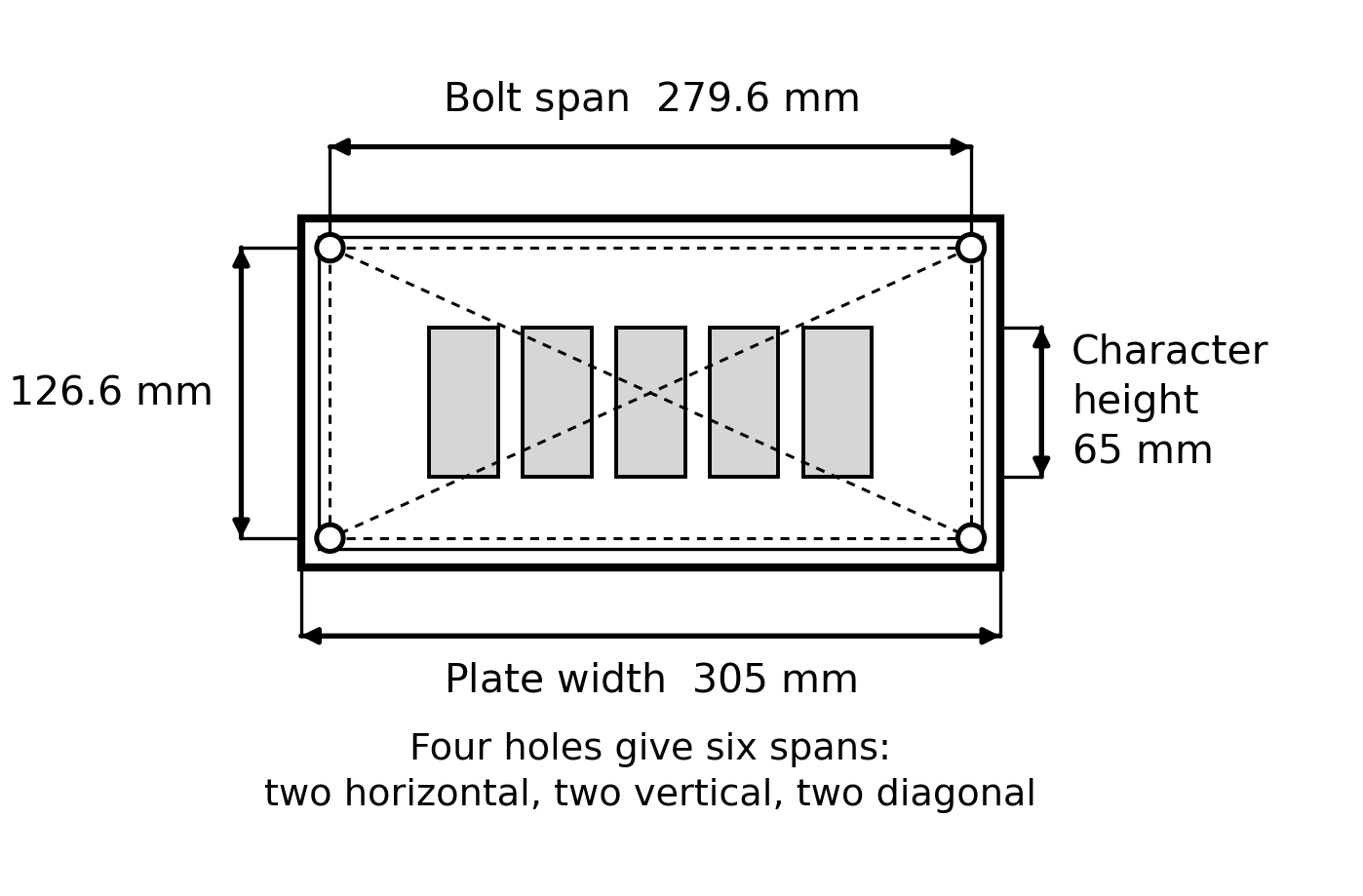}
\caption{Face-on geometry: Channel reference dimensions and the six mounting-hole spans\label{Figure:spans}}
\end{figure}

\begin{figure}[htbp]
\centering\includegraphics[width=\linewidth]{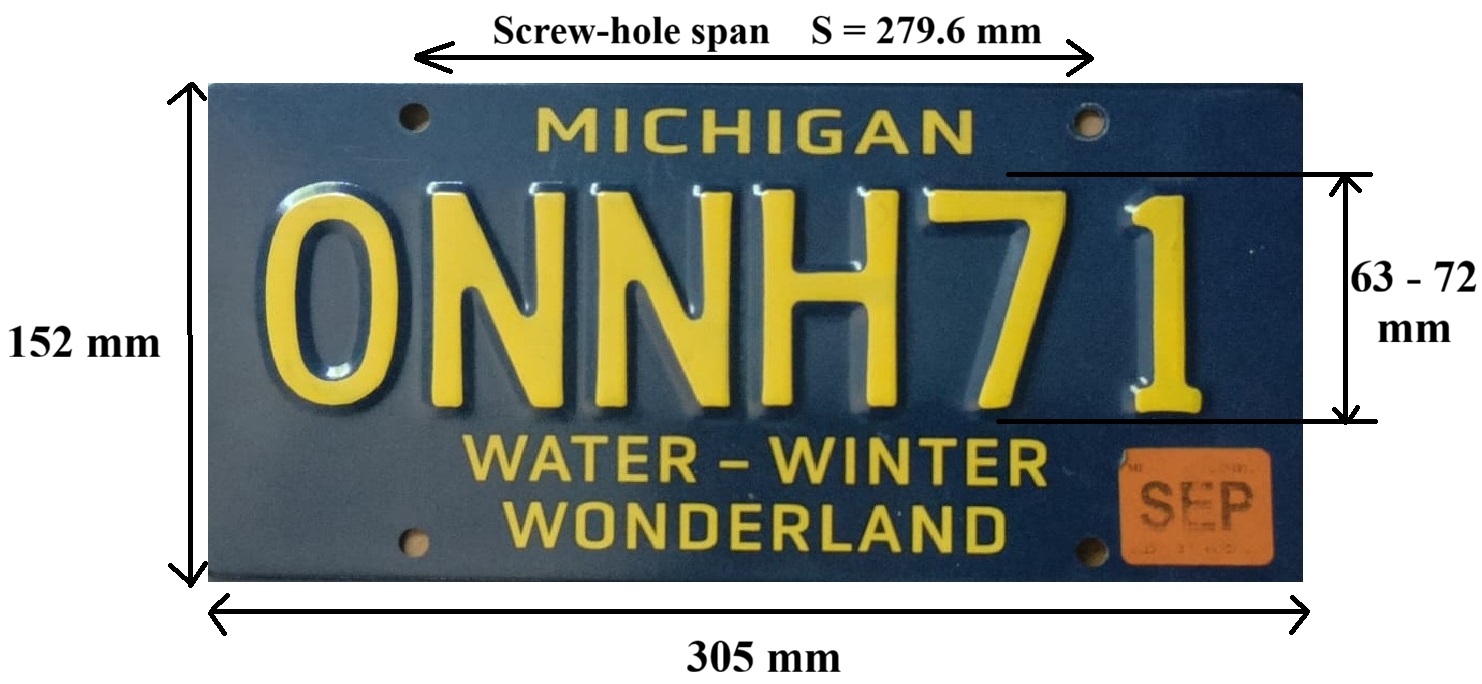}
\caption{The four standardized dimensions each ranging channel scales by, measured on an issued USA passenger plate \label{Figure:dims}}
\end{figure}

Figure~\ref{Figure:geom} shows both quantities on the measurement geometry, $D$ being the range to the plate and each rotation drawn about the plate axis it turns: vertical tilt $\theta$ about $x_p$, horizontal tilt $\psi$ about $y_p$, and in-plane roll $\phi$ about the plate normal $z_p$. Figure~\ref{Figure:spans} names the standardized dimensions against which the three channels scale and the six spans provided by the four mounting holes, and Fig.~\ref{Figure:dims} carries the same dimensions on a plate issued instead of a drawing, since the argument of this paper is based on those dimensions being real. Distance and angle are read from different properties of the same four points. Distance follows from apparent size, a known length subtending fewer pixels as it recedes, which is Eq.~\eqref{eqn:pinhole}. Angle follows from apparent shape: a rectangle seen square-on images as a rectangle, while one seen obliquely images as a trapezoid whose near edge is longer than its far edge, and it is that departure from a scaled rectangle rather than any single measured length that fixes the attitude. One consequence is that the angles need no metric dimension at all, only the knowledge that the plate is a rectangle of standard aspect ratio, whereas the distance needs the physical size in meters.

The outline is the method's most reliable ruler, for the reason Eq.~\eqref{eqn:sizelaw} predicts. Being the largest single feature, a fixed corner-localization error is the smallest fraction of its imaged extent. It is also recovered from four high-contrast corners, which the detector fixes cleanly across almost the whole envelope. Its main practical hazard is an aftermarket frame or dealer surround covering the printed rim, in which case the outline localizes to the visible plate edge and retains the small residual offset the calibration of Sec.~\ref{sec:system} removes. Because the channel never needs a character to be read, it still works on dirty, stylized or foreign-language plates whose serial cannot be recognized.

\subsection{The Mounting-Hole Channel and Fixture Geometry}
The third distance comes from the plate's mounting holes and depends on neither the text nor the outline. The American Association of Motor Vehicle Administrators (AAMVA) standard puts four bolt holes at a fixed inset of $12.7$~mm from the plate edges, giving a horizontal center-to-center span of $279.6$~mm and a vertical one of $126.6$~mm. A Hough circle search~\cite{duda1972use} finds the holes, steered to their expected positions by the recovered pose, and an imaged separation inverts through Eq.~\eqref{eqn:pinholecorr}.

Four holes define six center-to-center spans and not one, and the channel uses every span the mounting leaves visible: two horizontal, along the upper and lower hole rows; two vertical, along the left and right columns; and two diagonal. Writing the hole inset as $c = 12.7$~mm, the three distinct true lengths and the plate-axis directions along which they lie are the following.
\begin{equation}\label{eqn:spans}
\begin{aligned}
s_{\mathrm{h}} &= W - 2c = 279.6~\text{mm}, &\quad (d_x,d_y) &= (1,0)\\
s_{\mathrm{v}} &= H - 2c = 126.6~\text{mm}, &\quad (d_x,d_y) &= (0,1)\\
s_{\mathrm{d}} &= \sqrt{s_{\mathrm{h}}^2+s_{\mathrm{v}}^2} = 306.9~\text{mm}, &\quad (d_x,d_y) &= (\pm s_{\mathrm{h}}\, s_{\mathrm{v}})
\end{aligned}
\end{equation}

Each span is divided by the foreshortening factor of its own direction, obtained from $\mathbf{d} = d_x\mathbf{u} + d_y\mathbf{v}$ as in Sec.~\ref{sec:pose}, before Eq.~\eqref{eqn:pinholecorr} is inverted; one shared factor would be wrong for four of the six. Their distances combine by the median, not the mean, since a single mislocalized hole center corrupts three of the six spans and the median rides that out where an average would not. Restricting the channel to the upper pair, as earlier work did, throws away two thirds of the geometry the standard guarantees and makes the estimate turn on two points instead of four.

\begin{figure}[htbp]
\centering\includegraphics[width=\linewidth]{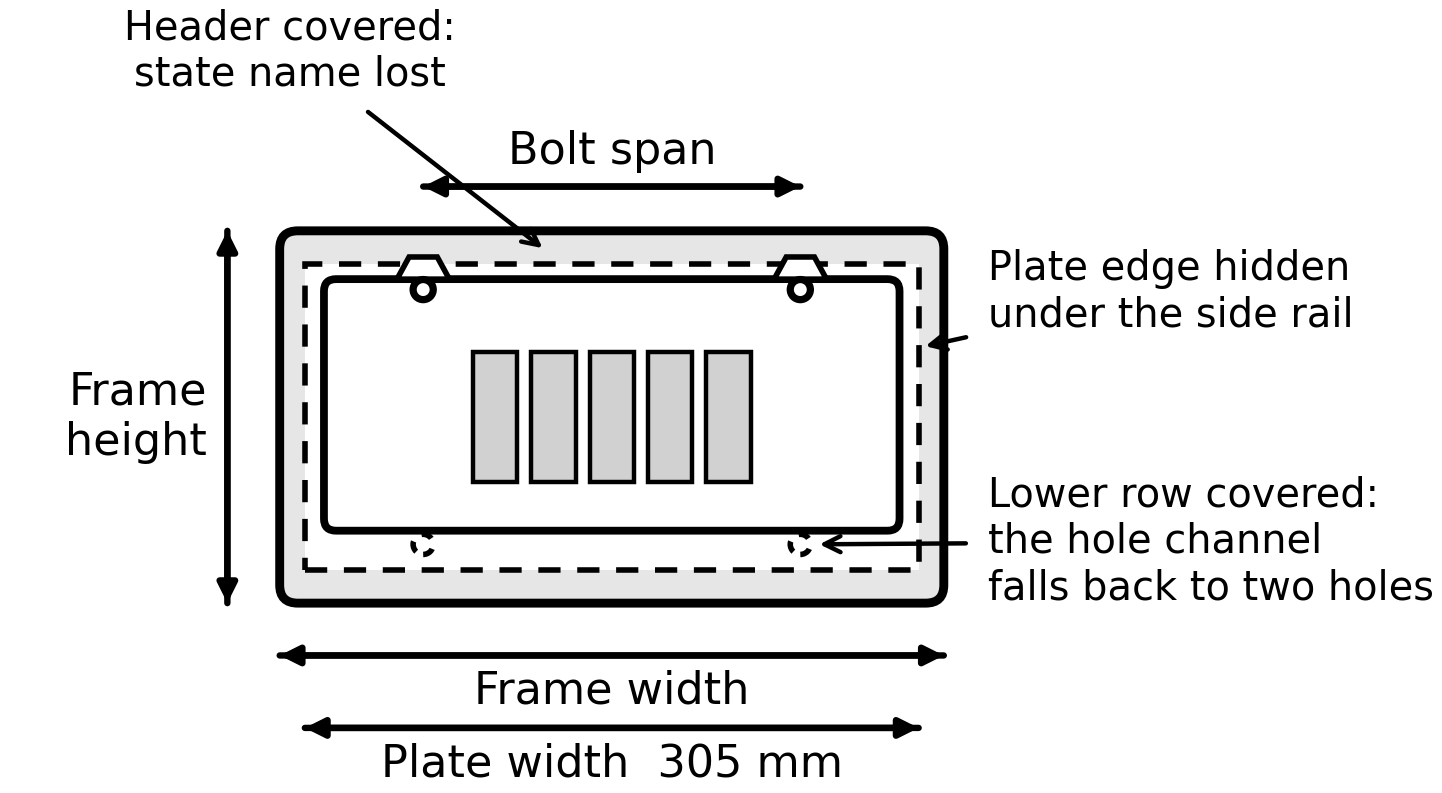}
\caption{What an aftermarket surround removes: the plate's side edges, the lower hole row, and the header carrying the state name\label{Figure:framed}}
\end{figure}

\begin{figure}[htbp]
\centering\includegraphics[width=\linewidth]{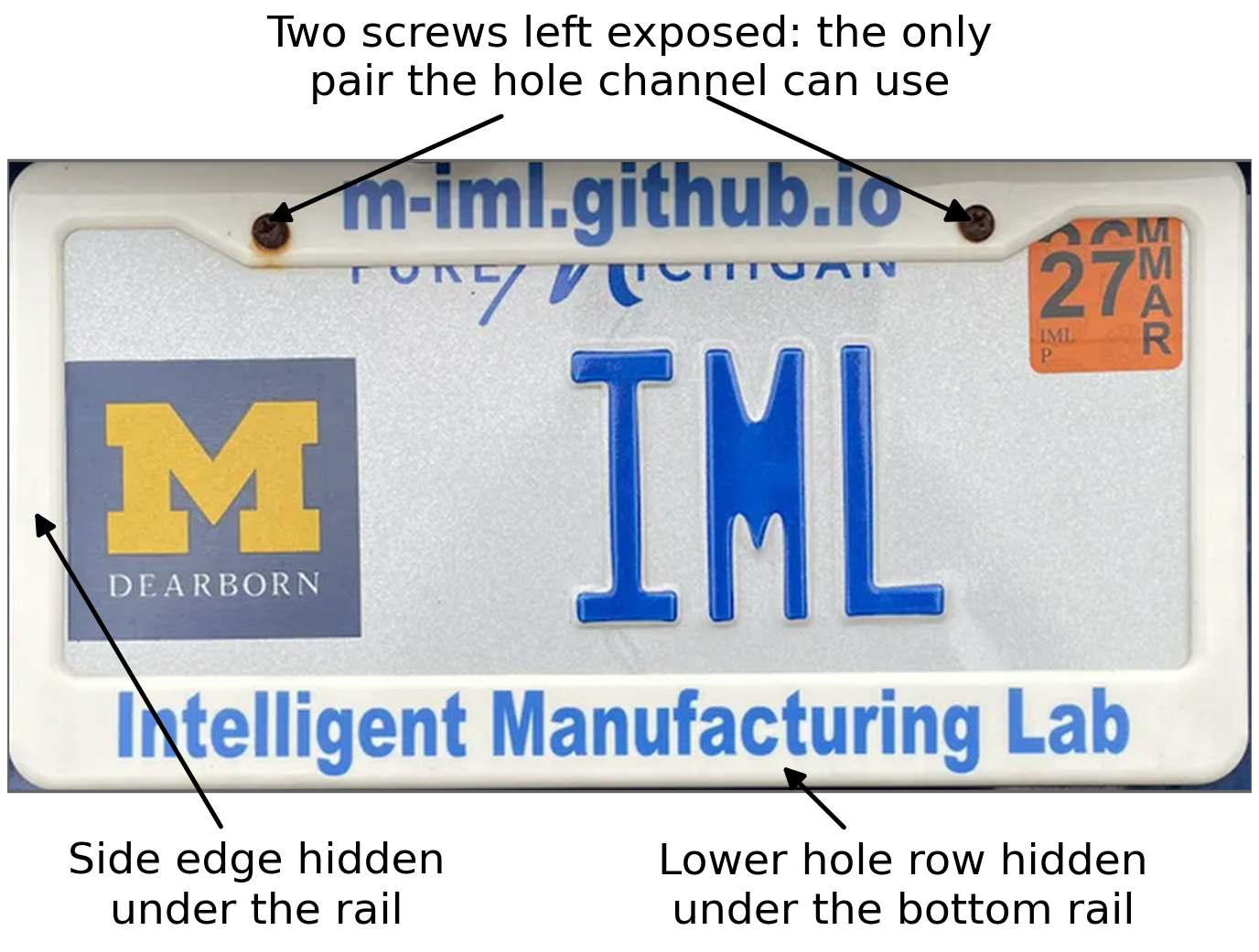}
\caption{The same obstruction on a real plate, with the two screws the frame leaves exposed and the two features it hides marked\label{Figure:framedphoto}}
\end{figure}

Admissibility is decided per frame rather than assumed, and that is what lets one rule cover both mounting cases. A plate bolted straight to the body exposes all four holes and yields all six spans. An aftermarket frame or dealer surround covers the plate perimeter, and the detector already reports the obscured perimeter edges; mapping an obscured edge to the holes it hides marks those holes unusable, and whatever spans remain are used. A frame covering the lower edge therefore reduces the channel to the upper row alone, the two-hole case, but it arrives there by measurement instead of being hard-coded to it, and a frame covering only the left edge leaves the right-hand vertical span available where a fixed upper-pair rule would have had nothing. Figure~\ref{Figure:framed} draws that case and Fig.~\ref{Figure:framedphoto} shows it on a plate wearing an aftermarket surround. Three things go at once. The side rails cover the plate's own left and right edges, so the outline channel localizes to the visible aperture and not to the printed rim; the lower rail buries the bottom hole row, leaving the two screws the frame notches its top rail to clear; and the header carrying the state name disappears, which removes the strongest of the four jurisdiction readers and sends the character channel to its default height. What survives is one horizontal span between two visible holes, and an outline whose edges are the frame's rather than the plate's. Both remain measurements, and the per-hole flag and the widened variance of this Section are what stop the fusion from treating either as though it were the unobstructed case.

Where a hole is neither visible nor recoverable, the channel falls back on the pose-projected bolt pattern, mapping the known AAMVA positions into the image through the solved pose so that a span is still produced when every hole is hidden. A per-hole flag records which centers were measured and which projected, and the channel's variance is widened whenever it leans on the projection, a projected center inheriting the pose solve's error rather than the image's.

Imaged hole centers sit a little inside the nominal bolt pattern, so the span reads slightly short and the distance slightly long, a bias the range-dependent calibration of Sec.~\ref{sec:system} removes. Tied to neither the serial nor a complete outline, this is the channel most likely way to survive when a frame hides part of the rim and grime hides the text, which is why the fusion still returns a reading in cases that would defeat any single cue.

\section{The MetroSim Metrology-Exact Rendering Testbed}\label{sec:sim}
\subsection{Testbed Setup}

Validating a ranging method to the $1\%$ level is hard on the road. The true distance to each plate has to be surveyed, as it is for extrinsic calibration to a physical target~\cite{cao2026extrinsic}. Presenting the same plate at a controlled sequence of distances and angles, so that one factor can be separated from another, is not easy either; and the space of jurisdictions, viewing geometries, and lighting conditions is far larger than any feasible campaign could cover reproducibly. We therefore validate the system entirely in software with MetroSim, a metrology-exact rendering testbed, following what is now standard practice in using controlled synthetic scenarios to develop and test driving perception~\cite{dosovitskiy2017carla,janai2020computer}. What distinguishes MetroSim is a ground truth exact by construction: the plate is drawn at a known physical size and projected by the same geometry the method later inverts, so the commanded distance is the true distance, known to machine precision instead of measured with a tape. Residual error is attributable to the ranging pipeline alone and not to an uncertain reference. MetroSim is a metrology instrument first and a renderer second, photorealism being sacrificed wherever it would compromise the exactness of a measured dimension.

For a chosen jurisdiction the framework draws a plate at that state's regulated character height with the FHWA stroke-to-height and spacing ratios, its standardized $305$~mm outline, its embossed rim and corner registration sticker, and mounting holes at the AAMVA insets, so every feature the pipeline measures has a known true value. The synthesized plate is then projected into the frame by Eq.~\eqref{eqn:pinhole} at a commanded distance and pose through a full perspective homography, so oblique views show true foreshortening, not a mere in-plane rotation and the pose-correction stage is exercised as it would be in the field. The camera modeled throughout is an automotive module built on a $2880\times1860$ sensor with a $38^{\circ}$ horizontal field of view, giving $f\approx3967$~px at full width, though the focal length is a parameter and other sensors are configured by editing a single profile.

Two modes stress different parts of the system. A controlled pose study places a single plate at a commanded distance and viewing angle and sweeps the two independently, isolating the effect of each factor on each channel, which is the basis for the per-channel characterization. A continuous driving scenario renders an approaching lead vehicle from a bumper-mounted ego camera on a perspective road, with the plate mounted on a vehicle body of realistic color and shading and the scene including ego roll and lateral sway, defocus, and speed-dependent motion blur. Adverse illumination is applied after the plate is composited, so the detector and temporal filter meet motion and degraded light as a real sensor would. The driving mode also exercises the innovation gate and the closing-speed estimate, which the static pose study does not. Both modes emit the exact ground-truth distance for every frame, so both feed the same scoring.

The validation sweeps all 51 USA states, exercising the full spread of regulated character heights and plate designs. Distance is drawn continuously and uniformly from $1$ to $12$~m instead of at a handful of fixed stops, which characterizes the error evenly across the whole range and keeps any artifact of favorably chosen distances out of the estimate-versus-truth relation. Yaw reaches $30^{\circ}$ and pitch $20^{\circ}$, the obliquities a bumper-mounted camera meets as a lead vehicle changes lane or crests a grade. Five illumination conditions overcast, dusk, night, rain, and glare are applied as a separate robustness probe. For every frame we record whether the plate was detected, counting a box that latches onto the background or a single glyph as a miss rather than a spurious success, and for each channel and for the fusion the absolute percentage error $|d_i-D|/D$. Plate pose is scored by the absolute error between commanded and recovered in-plane roll. Since MetroSim needs no instrumented test track and no surveyed distances, the entire envelope is exercised reproducibly in software, and any change to the system a new detector guard is built and a revised fusion rule is re-validated across the whole envelope before a vehicle is ever involved.

Four families of measure are computed over a sweep of $N$ samples, each characterizing a different aspect of the error. Accuracy is summarized by the mean absolute percentage error (MAPE) and its distribution tail
\begin{equation}\label{eqn:mape}
\begin{gathered}
\mathrm{MAPE} = \frac{100}{N}\sum_{k=1}^{N}\frac{|\hat{D}_k - D_k|}{D_k},\\[2pt]
\mathrm{P}q = \operatorname{percentile}_q\bigl[\,|\hat{D}_k - D_k|\,\bigr]
\end{gathered}
\end{equation}

Percentiles are reported because a warning function is endangered by its worst frames, not its average one. Since the estimator also reports a variance, that claim is tested rather than assumed: with the normalized residual $z_k = (\hat{D}_k - D_k)/\hat{\sigma}_k$, a correctly calibrated estimator has an average normalized estimation error squared (ANEES) of unity and a coverage set by the normal law.
\begin{equation}\label{eqn:anees}
\mathrm{ANEES} = \frac{1}{N}\sum_{k=1}^{N} z_k^2 = 1,\qquad
\Pr\bigl(|z| \le \kappa\bigr) = \operatorname{erf}\!\left(\kappa/\sqrt{2}\right)
\end{equation}

A correctly calibrated estimator places $68.3\%$ and $95.4\%$ of samples within one and two standard deviations. A value above unity means the reported $\hat{\sigma}$ is too small and the estimator is over-confident, the unsafe direction for any filter consuming it. Systematic error is separated from random error by regressing the estimate on the truth.
\begin{equation}\label{eqn:agree}
\hat{D} = a D + b + \varepsilon
\end{equation}
A slope $a\neq 1$ betrays a scale error a wrong plate constant or an uncorrected foreshortening while an intercept $b\neq 0$ betrays a constant offset, two faults a single error average would not distinguish. Finally, since a distance error matters only through the alert it changes, the issued warning is judged by the warning the same rule returns from the exact simulated state, and the two asymmetric rates are reported separately: the missed-alert rate over frames whose true state warrants an alert, and the false-alert rate over frames whose true state does not.

Table~\ref{tab:constants} lists every constant the method uses together with its provenance, because a reader is entitled to separate the numbers fixed by something outside the work from the ones chosen here. Three are set by a published standard and are not free parameters at all: the plate outline and bolt span come from the AAMVA specification, and the two warning times from the National Highway Traffic Safety Administration (NHTSA) forward collision warning criteria. Six were measured, from the error model or from the sweep itself, and are reported with the measurement that fixed them. The remaining four were chosen on engineering grounds.

The innovation gate at five standard deviations, the six-frame re-initialization count, and the velocity dead band were set from the physical reasoning given above and then left alone; none was tuned on the reported results, and none was swept as the consensus threshold. Their sensitivity is therefore unmeasured, a weaker statement than the consensus gate can make. The one thing to be said for them is that they are not fitted to the numbers this paper reports, so they are not a route by which the test set could have leaked into the method.

\begin{table}[htbp]
\caption{Design constants and their provenance\label{tab:constants}}
\centering\small
\begin{tabular}{@{}p{0.24\textwidth}p{0.18\textwidth}p{0.50\textwidth}@{}}
\hline
Constant & Value & Provenance \\
\hline
Plate outline width      & $305$~mm        & AAMVA standard \\
Bolt-hole span           & $279.6$~mm       & AAMVA standard \\
Character height         & $63$ - $72$~mm & State regulation, FHWA series \\
Caution time             & $2.4$~s          & NHTSA warning criterion \\
Danger time              & $1.5$~s          & NHTSA imminent-threat criterion \\
\hline
Character prior $\sigma$ & $0.153D$         & Error propagation from feature size \\
Outline prior $\sigma$   & $0.054D$         & Measured, $E|e| = \sigma\sqrt{2/\pi}$ \\
Hole prior $\sigma$      & $0.088D$         & Measured, as above \\
Hole calibration gain    & $0.9992 + 0.00742D$ & Fitted on half the sweep, scored on the held-out half \\
Measurement CV           & $0.06$           & Measured error of the fused distance \\
Consensus gate           & $0.40$           & $1.3\sigma$ of character-channel spread; swept over $0.05$ - $\infty$ \\
\hline
Acceleration disturbance & $4$~m/s$^2$      & Chosen: ordinary braking \\
Innovation gate          & $5\sigma$, $20$~m/s & Chosen: unmeasured sensitivity \\
Reinitialization count   & $6$ frames       & Chosen: unmeasured sensitivity \\
Velocity deadband        & $0.08$~m/s       & Chosen: corner jitter at close range \\
\hline
\end{tabular}
\end{table}

\subsection{Other Cameras and Vehicle Classes}\label{sec:reconfig}
Two properties carry the method and MetroSim well beyond the single sensor and vehicle class reported here. That camera enters the mathematics only through the focal length $f$ and the working image width, so a different sensor, whatever its resolution, pixel pitch or field of view is accommodated by editing one profile block. That block records the working width and either a calibrated $f$ or the datasheet horizontal field of view from which $f=(W/2)\cot(\theta_H/2)$ follows. Because $f$ scales with the width at which frames are actually captured, a profile calibrated at one resolution transfers to another by that ratio alone. Nothing in the detection, fusion or filtering code changes. The same source therefore runs unaltered on a narrow-field forawrd camera for highway following or a wide field module for urban ranging and MetroSim re-derives exact ground truth for whichever profile is loaded, so a new camera is characterized across the full envelope in software before it is ever mounted.

Every channel measures the plate and not the vehicle, which leaves the method indifferent to the body the plate is bolted to. The $305$~mm outline, the $279.6$~mm hole span and the regulated character height are identical on a sedan, an SUV, a pickup, a van or a heavy truck, so no vehicle model, body style or width has to be recognized before he distance can be produced. Only the plate's mounting height never enters the distance computation, since it moves the plate;s position in the image without altering the size it subtends, and size is what the channel measure. Extending coverage to a jurisdiction or vehicle market with a differently sized plate requires only substituting that standard's outline width, hole span and character height, the three constants the channel divide by leaving the pipeline otherwise intact.

\begin{figure}[htbp]
\centering\includegraphics[width=\textwidth]{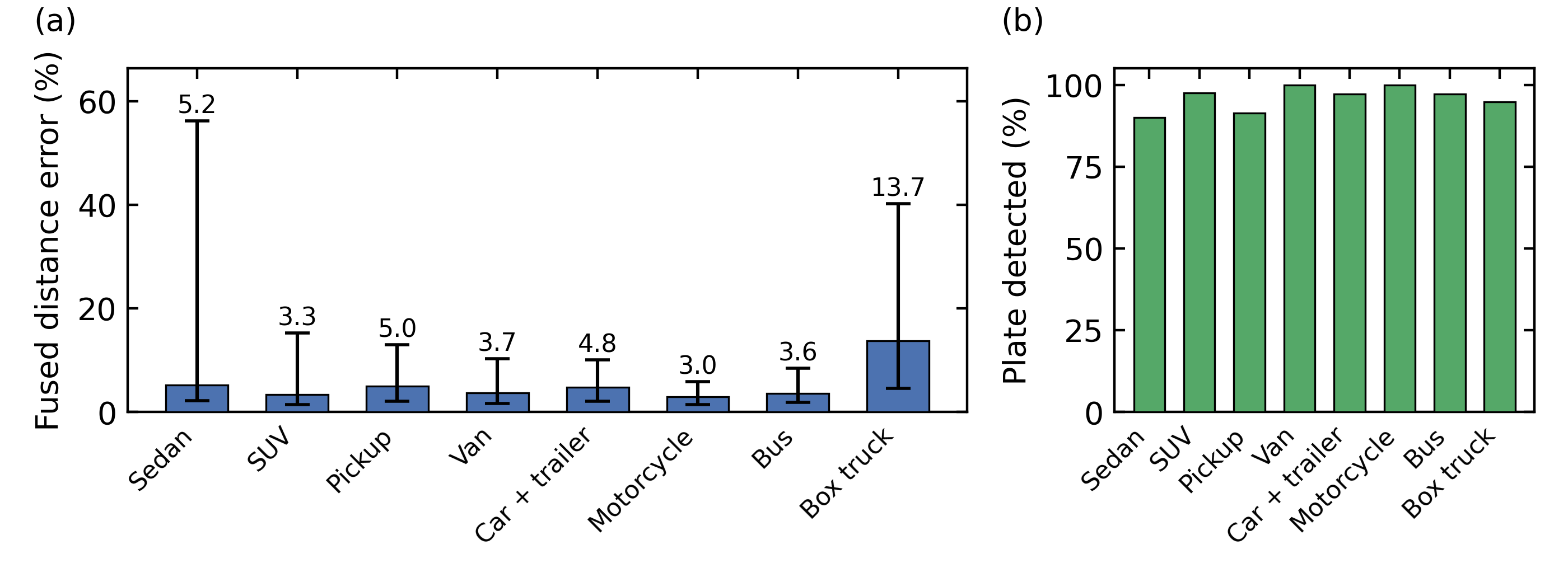}
\caption{Fused ranging error and detection rate across eight vehicle classes\label{Figure:vehicles}}
\end{figure}

The driving scenario exercises the second property directly, mounting the plate on bodies drawn from several vehicle classes of differing width, height, and plate elevation, and the ranging accuracy is unchanged across them because the metric reference is the plate, not the car. Figure~\ref{Figure:vehicles} shows the test of eight vehicle classes in the driving scenario. Median fused error stays between $3.0\%$ and $5.2\%$ for seven of them and the plate is found on $90\%$ to $100\%$ of frames, with the box truck the one outlier at $13.7\%$ and its plate sits lowest and is the most often occluded by the vehicle's own structure. The upper quartiles for the sedan and the box truck are nonetheless much wider than their medians, and that spread is the false-lock failure of Sec.~\ref{sec:conclusion}, not a loss of ranging precision. This evidence has one limit. Every class is rendered carrying the standard passenger plate, so the figure demonstrates indifference to mounting position and vehicle geometry, not to a smaller plate such as those fitted to motorcycles in many jurisdictions.

\section{Results and Discussion}\label{sec:results}

\subsection{Detection Reliability and Ranging Accuracy}\label{sec:resdet}
With the localization achieved in Sec.~\ref{sec:system}, the plate was localized in every frame across the entire sweep of jurisdictions, distances, and angles, and no jurisdiction failed to be detected in any pose. Such completeness is not automatic. At the far end of the range the plate spans only a hundred or so pixels, and a naive detector tends either to miss it in the background or to collapse its box onto the glyph cluster of a sparse plate; the character-anchored recovery holds the plate when its outline is low-contrast with the vehicle, and the trim guard prevents the collapse. Because detection does not fail within the tested envelope, the ranging errors below are true ranging errors, not detection artifacts, which is the separation a controlled study is meant to provide.

This figure is a true-positive rate. Every frame in the sweep contains a plate, so it can report how often a plate present is found and can say nothing about how often a plate absent is reported. The distinction is not academic. The detector selected the highest-scoring candidate with no absolute threshold, so on a frame containing no plate it returns whatever structure scored highest, which on plate-free clutter occurred on $87.5\%$ of frames. No rendered scene can expose this, containing as it does a plate on a plain background so that the highest-scoring candidate is the plate whatever the threshold. We now require a candidate to exceed a floor measured from the two score distributions. On plate-free clutter the winning score has a median of $0.713$, whereas plates score $0.759$ at minimum and $1.100$ at the median, so a floor of $0.750$ rejects $79.3\%$ of the clutter while losing no plates. The floor costs nothing on the sweep, where detection stays complete, and the false-positive rate that matters is the one a road trial would measure.

\begin{figure}[htbp]
\centering\includegraphics[width=\linewidth]{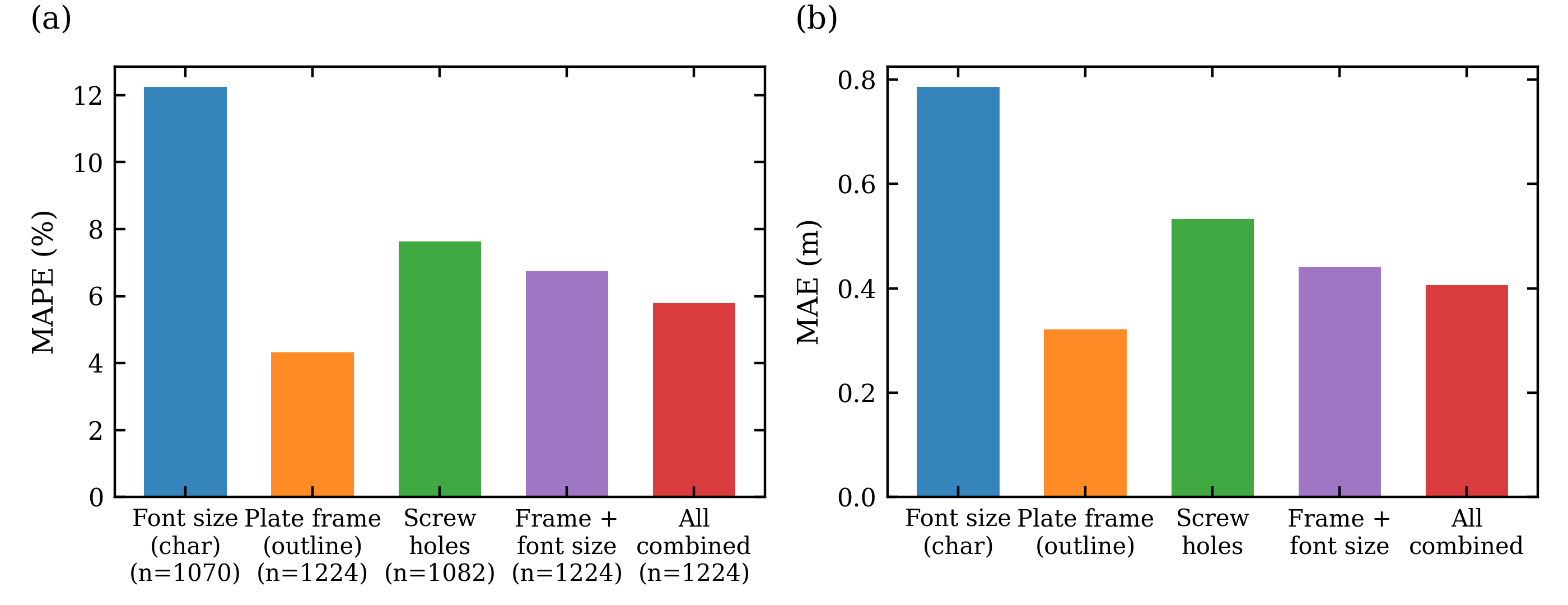}
\caption{Ranging accuracy by channel over the continuous sweep, as (a) the mean absolute percentage error and (b) the mean absolute error in meters\label{Figure:ablation}}
\end{figure}

The central result is that the accurate channels are the coarse ones. Table~\ref{tab:channels} and Fig.~\ref{Figure:ablation} compare the MAPE of each channel over the full sweep. Most accurate is the plate outline at $4.32\%$, with the mounting-hole span behind it at $7.63\%$; both rest on large, stable geometric distances the detector recovers cleanly. Character height however, is the least accurate overall at $12.24\%$ despite being the most precise feature at close range, and Eq.~\eqref{eqn:sizelaw} explains why. Glyphs are the smallest of the three features, roughly a fifth the width of the plate outline, so a fixed segmentation error of a pixel or two is a much larger fraction of its imaged height, and that fraction grows as the plate recedes. Once the object is small in the image, the finest ruler is not the best ruler.

Fusing the calibrated channels by Eq.~\eqref{eqn:fuse} gives $5.79\%$ with a median of $2.17\%$, which is no improvement on the outline channel alone and we report it as such. That follows from the evidence policy, not from the weighting, and it separates cleanly. On the $78.4\%$ of frames where a trustworthy attitude is recovered, the fusion returns $3.67\%$ where the outline return $3.13\%$, a margin of $0.5\%$ point. On the rest the system narrows to character height by the rule of Sec.~\ref{sec:system}, and there the fusion returns $13.48\%$ where the outline return $8.64\%$, the channel it discards being the better one in that setting. A fifth of the frames thus contribute about half the total error. This is a deliberate trade and not a defect in the fusion. The rule exists for the field case in which the outline is lost to low contrast, a case the controlled sweep does not present, and the alternative settings cost $4.57\%$ with no narrowing and $5.15\%$ when the outline is retained beside character height. The fusion's actual benefit is unchanged, and it is robustness, not accuracy. Size-scaled variances hold the fragile character channel down, the consensus gate removes a corrupted reading before it can move the estimate, and the evidence gate keeps a channel with no measurement behind it from voting at all. In this sweep fusion is a mechanism for not being wrong in the ways a single channel can be, and the outline alone is the more accurate estimator when it is available.

\begin{table}[htbp]
\caption{Ranging accuracy by channel over the full sweep\label{tab:channels}}
\centering{%
\begin{tabular}{l r}
\toprule
Ranging channel & MAPE [\%] \\
\midrule
Character height    & 12.24 \\
Plate outline ($305$~mm)          &  4.32 \\
Mounting-hole span ($279.6$~mm)   &  7.63 \\
Consensus-gated fusion            &  5.79 \\
\bottomrule
\end{tabular}
}%
\end{table}

\begin{figure}[htbp]
\centering\includegraphics[width=\linewidth]{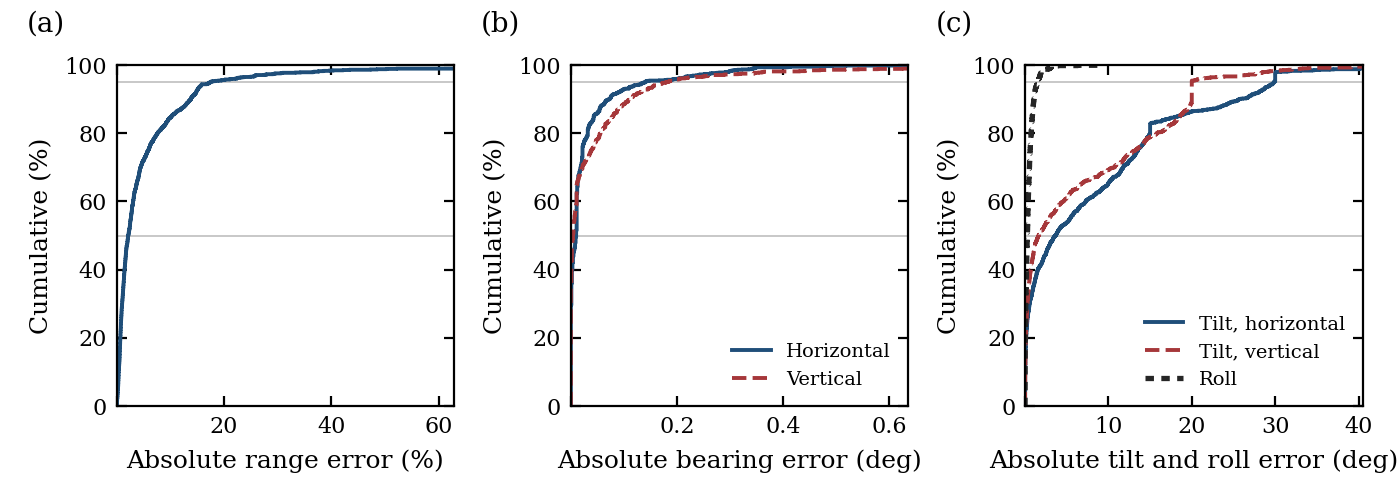}
\caption{Cumulative distribution of the absolute error in each reported output, for (a) the range, (b) the two bearings and (c) the two tilts and the roll, each panel on its own axis because the outputs span three orders of magnitude\label{Figure:outbox}}
\end{figure}

\begin{figure}[htbp]
\centering\includegraphics[width=\linewidth]{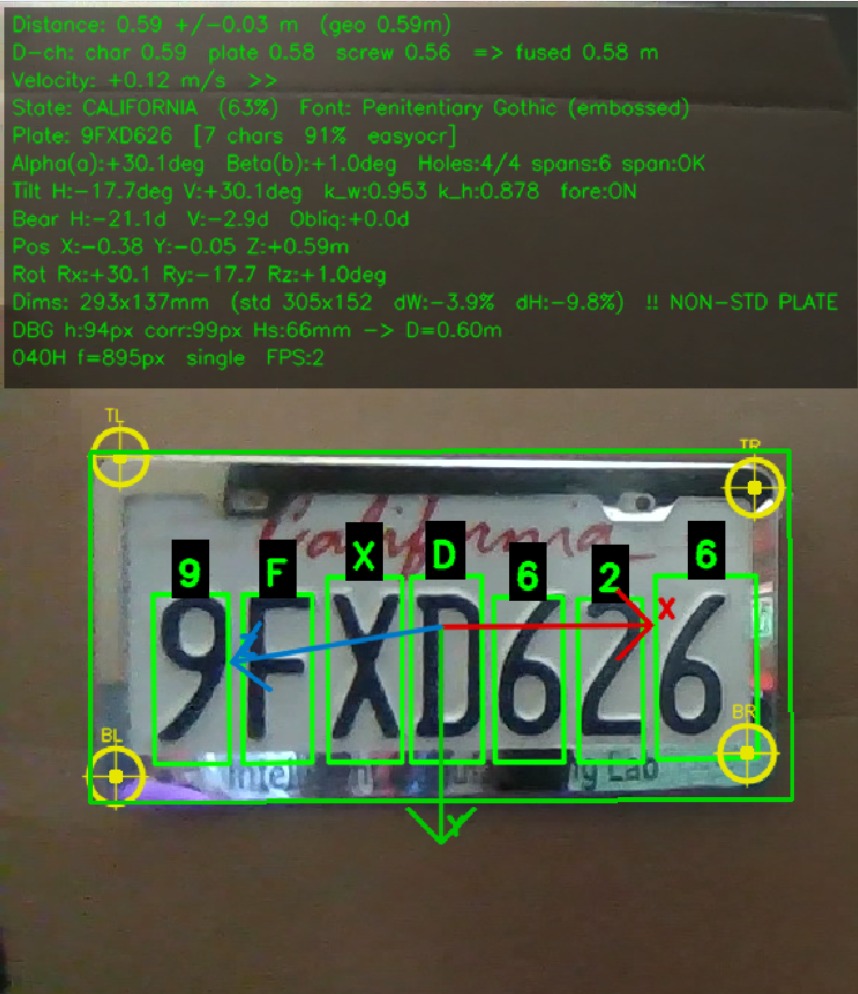}
\caption{A single live camera frame carrying the three ranging channels, the recovered plate axes, and the located mounting holes\label{Figure:hud}}
\end{figure}

\subsection{Algorithm Outputs}\label{sec:resout}
Five numbers per frame come out of the system, not one: three fixing where the plate is and two fixing which way it faces. Reporting only the range, as the accuracy literature on this method has done, characterizes a fifth of the output. Table~\ref{tab:outputs} scores all five on the same renders, Fig.~\ref{Figure:outbox} gives their cumulative error distributions, and Fig.~\ref{Figure:hud} shows one frame as the system delivers it, with the localized quadrilateral, the segmented serial, the recovered plate axes and the four mounting-hole positions drawn over the image that produced them. Each panel of Fig.~\ref{Figure:outbox} carries its own linear axis, since the outputs span three orders of magnitude, and the step at the origin of panel (b) is the bearing error recorded as exactly zero.

Two of the six rotational and translational quantities the pose solve produces are therefore reported and one is not, an exception easy to read as an omission. A rigid body has six degrees of freedom, three of translation and three of rotation, and the plate is a rigid body. But a plate is an oriented rectangle, not a ray. Which way it faces is a direction, and a direction in three dimensions takes two numbers, while the third rotational degree of freedom is the spin about that facing, the in-plane roll. Roll does not change the collision geometry and so is not part of the answer. Roll is nonetheless computed on every frame and cannot be dropped from the calculation, the foreshortening factors of Eq.~\eqref{eqn:kfactor} being written on the full rotation, and the roll sensitivity quantified in Sec.~\ref{sec:pose} would pass directly into the outline and character distances.

Position is expressed as a range and two bearings instead of $X$, $Y$, $Z$. Both forms hold identical information, since $X = D\sin b_h$ and so on, but a bearing error is an angle whose meaning does not change with distance whereas an error in $X$ is a length that grows with it, so only the bearing form can be summarized in a single row. Roll is listed although it is not among the five, because it is computed, because every foreshortening factor depends on it, and because it is the pose quantity this system recovers best, so omitting it would hide the one strong pose result.

Three features of the table stand out. The bearings are the most accurate output by two orders of magnitude, at a mean error of $0.03$ - $0.04^{\circ}$, a bearing being read from where the plate's centroid falls in the image and needing no scale, no pose and no metric constant it is the one output inheriting none of the difficulties the rest of the paper is about. The two tilts are the weakest at $8.4^{\circ}$ and $6.8^{\circ}$ mean, for the reason established in Sec.~\ref{sec:pose}. And the range distribution is severely tail-heavy: a median of $2.18\%$ beside a mean of $5.79\%$ and a maximum of $370\%$. That single worst frame is a mis-detection, not a ranging error, and it is why the median is quoted beside every mean in this paper and why the temporal filter includes an innovation gate a $370\%$ error reaching a warning threshold would be a false alert of the most damaging kind.

\begin{table}[htbp]
\caption{Absolute error in each of the five reported outputs and in roll over the controlled sweep, as a percentage for the range and in degrees for the bearings and angles\label{tab:outputs}}
\centering\small
\begin{tabular}{@{}lrrrrr@{}}
\hline
output & min & mean & median & P95 & max \\
\hline
range $D$ (\%)              & 0.00 & 5.79 & 2.18 & 17.41 & 370.33 \\
bearing, horizontal ($^{\circ}$) & 0.00 & 0.03 & 0.01 & 0.14 & 0.72 \\
bearing, vertical ($^{\circ}$)   & 0.00 & 0.04 & 0.00 & 0.18 & 1.38 \\
tilt, horizontal $\psi$ ($^{\circ}$) & 0.00 & 8.36 & 3.66 & 29.87 & 75.40 \\
tilt, vertical $\theta$ ($^{\circ}$) & 0.00 & 6.82 & 1.62 & 20.00 & 61.34 \\
roll $\phi$, computed ($^{\circ}$)   & 0.00 & 0.44 & 0.27 & 1.56 & 8.75 \\
\hline
\end{tabular}
\end{table}

\subsection{Behavior on the Camera}\label{sec:bench}
Every number above comes from rendered frames, so we also ran the same pipeline on the sensor it was written for. Figure~\ref{Figure:hud} is a single live frame at $1280\times720$ from the automotive module of Sec.~\ref{sec:sim}, with a Michigan plate held in an aftermarket frame at roughly two thirds of a meter. It demonstrates behavior, not accuracy, a hand-held plate having no surveyed distance and the frame therefore no ground truth for scoring. The demonstration is nonetheless worth making, since several of the mechanisms this paper argues for are visible acting together in one frame and none of them can be seen in a table of averages.

Four of those mechanisms are legible on the overlay. The three ranging channels report $0.69$, $0.64$ and $0.68$~m and fuse to $0.66$~m, agreeing to within $0.05$~m without any of them having been told what the others found. The frame is detected, and the hole channel responds as Fig.~\ref{Figure:flowchan} draws it: four centers are located, the covered ones excluded, and the channel falls back to the single top-row span instead of fabricating the six it would use on a bare plate. The recovered attitude of $9.5^{\circ}$ horizontal and $3.0^{\circ}$ vertical tilt yields foreshortening factors of $0.986$ and $0.999$, which is the geometry of Eq.~\eqref{eqn:kfactor} at work on a plate a reader can see is turned. The jurisdiction is identified as Michigan at $90\%$ confidence, selecting the $72$~mm character height instead of the $65$~mm default, and the serial is read. One caution belongs with the figure: the recovered outline of $314\times157$~mm sits $3.1\%$ and $3.5\%$ above the standard, and that agreement is not an independent check, the reported dimensions being the fused distance multiplied back through the same pinhole relation. It restates the channel agreement in millimeters instead of confirming it.

\begin{figure}[htbp]
\centering\includegraphics[width=\linewidth]{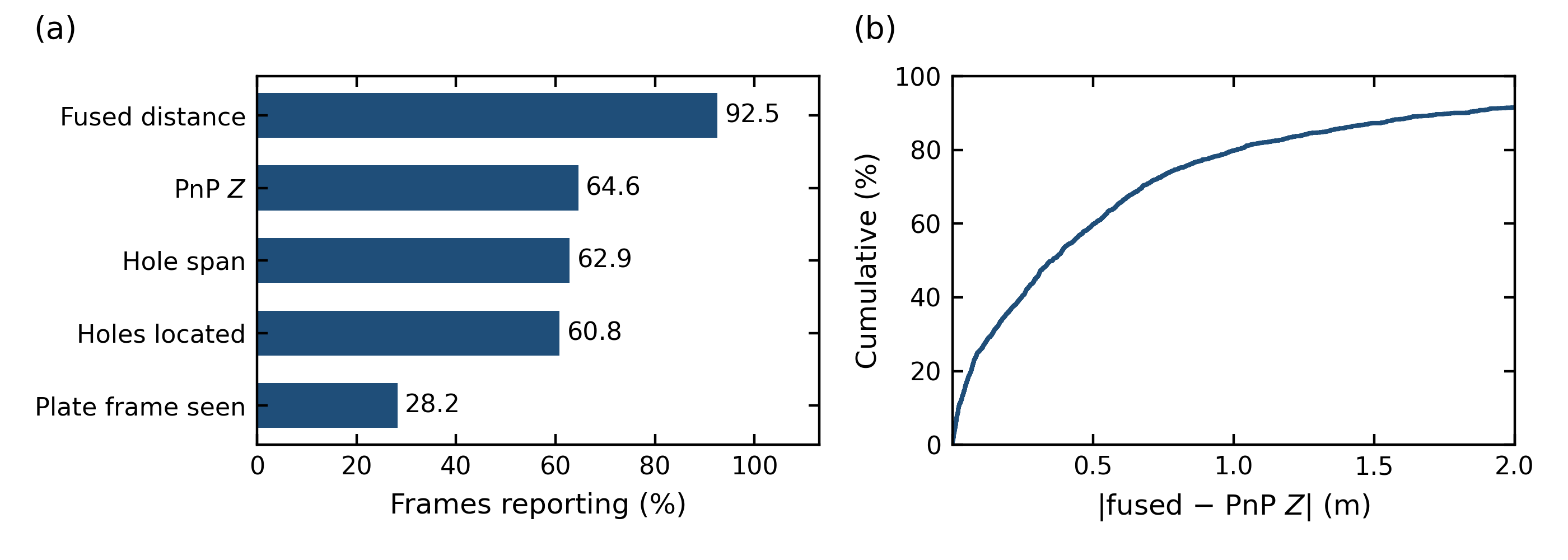}
\caption{(a) Per-channel availability and (b) Agreement between the fused distance and the pose solve \label{Figure:camera}}
\end{figure}

One frame shows mechanism and cannot bound behavior. Figure~\ref{Figure:camera} therefore scores the same pipeline over $3145$ frames from $66$ camera sessions, with the driving-scenario logs excluded so that nothing rendered counts as sensor data. Two things are measurable without a surveyed distance, and both bear on deployment. Availability is the first. The fused estimate reports on $92.5\%$ of frames, but the perspective-$n$-point solve reports on $64.6\%$ and the mounting holes are located on $60.8\%$, where the geometric channels reached $100\%$ in the rendered test run. This difference is the sim-to-real gap stated in the currency that matters to a fusion rule, and it is the strongest argument in this paper for carrying three channels rather than the most accurate one. Self-consistency is the second. The fused distance and the pose solve reach the same target through different features, and they agree to a median of $0.36$~m, with $60\%$ of frames within $0.5$~m and $80\%$ within $1$~m. This figure is consistency and not accuracy, exactly the evidence Sec.~\ref{sec:system} argues is insufficient in the prior work, and we claim no more from it here: two estimators sharing a detector can agree and still be wrong together. The absence of a surveyed reference on real hardware is the gap MetroSim exists to close, and these numbers describe its width rather than removing it.

\subsection{Dependence on Distance and Viewing Angle}

\begin{figure}[htbp]
\centering\includegraphics[width=\linewidth]{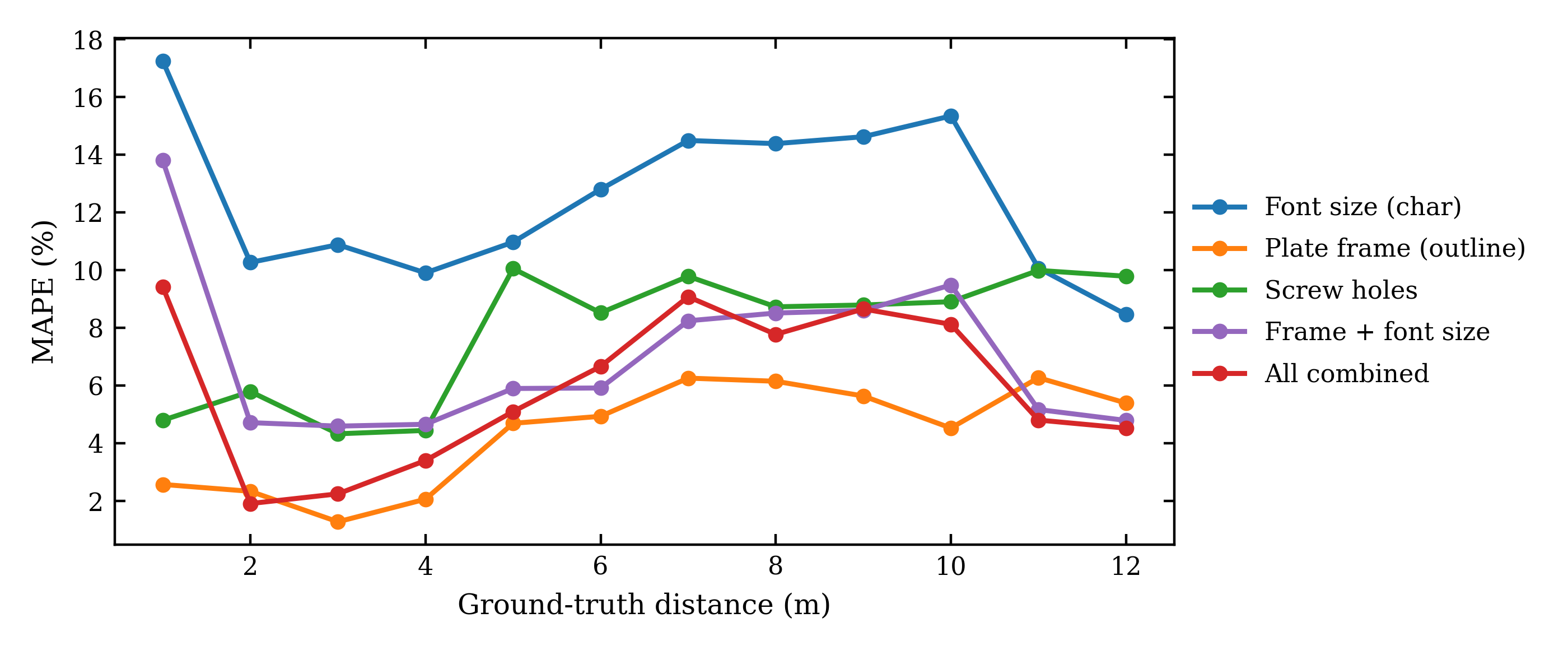}
\caption{Per-channel error against distance\label{Figure:vsdist}}
\end{figure}

Resolving the error by range, in Fig.~\ref{Figure:vsdist}, sharpens the per-channel picture and confirms the propagation analysis. All channels degrade with distance, as Eq.~\eqref{eqn:pinhole} demands, but at different rates. The character-height channel degrades fastest because its feature shrinks fastest in relative terms; by the far end of the range its error is several times the outline's. The outline and hole channels, whose features stay comparatively large, hold their accuracy far better, and the fusion tracks them at distance because the inverse-variance weighting shifts trust toward the channels whose variance grows more slowly. In practical terms character height earns its weight only at close range, while the geometric channels carry the estimate at the distances where a warning must be raised earliest an operating characteristic worth designing around instead of averaging away.

Oblique viewing foreshortens the plate, and uncorrected it biases every channel toward over-estimation by the amount Sec.~\ref{sec:pose} predicts. Measured on the outline channel before the correction of Eq.~\eqref{eqn:pinholecorr} was applied, the mean bias ran $-2.1\%$ frontally, $+3.4\%$ at $15^{\circ}$ of horizontal tilt and $+13.2\%$ at $30^{\circ}$; the increments of $3.5$ and $15.5\%$ points are $1/\cos\psi - 1$ to within a $0.1\%$ of a point, which identifies the mechanism beyond reasonable doubt. With the correction active the same measurements give $-1.2\%$, $+0.3\%$ and $+5.5\%$. The residual at $30^{\circ}$ is a shortcoming of the frames, not of the correction, on which the system declines to apply it, and separating the two makes that explicit. The quadrilateral refinement is accepted on $59.3\%$ of frames, at a rate flat with distance, not falling off at range. On those frames the recovered width foreshortening is $0.966$ at a true $15^{\circ}$ matching the ideal $0.966$, and the outline bias at $30^{\circ}$ is $+3.7\%$. On the frames where it is declined the same bias is $+8.9\%$, and the pooled $+5.5\%$ is a mixture of the two populations, not a property of either. The refinement also improves plain corner localization and not only obliquity: on frontal frames, where there is no foreshortening to remove, its acceptance cuts the mean outline bias from $-1.5\%$ to $-0.5\%$.
This choice deserves its own justification, because declining to correct looks like a failure to correct and is not. The asymmetry set out in Sec.~\ref{sec:pose} is what sets the policy: an uncorrected channel is wrong by a bounded amount, a wrongly corrected one by an unbounded amount. Declined frames are recorded instead of being silently treated as frontal, so their variance is widened and the fusion weights them accordingly.

A decorative or non-standard surround puts the angle at risk, since every estimator named so far reads the plate's border. The outline supplies both the corner-edge roll and the vanishing points, and a frame covering the border hides the mounting holes and removes the hole rows with them, so an unusual frame can in principle take the whole attitude with it. The serial cannot be reached that way, sitting inside the frame's aperture whatever the surround does. We therefore added it as a fourth estimator, with one requirement that turns out to decide how it can be used. It must be segmented in the raw frame instead of in the rectified warp where the pipeline already segments, because the warp is built from the plate corners and removes the very rotation the estimate would be measuring; glyph boxes taken there are horizontal by construction and would restate the corners instead of checking them. Segmented before rectification, the serial is a row of evenly spaced marks on a straight baseline, and the inclination of that row is the plate's roll. Only the location of the search region comes from the corners.

The value of that estimator was measured rather than assumed, and the measurement overturned our first reading of it. Judged on the ground truth alone, the serial is the weakest of the estimators, recovering roll to $1.27^{\circ}$ of mean absolute error where the corner angle reaches $0.41^{\circ}$, glyph centroids spanning a shorter baseline with noisier endpoints than the plate's long edges. From that we concluded it should not vote. The full sweep disagreed, and the sweep was right, admitting it moving the reported roll from $0.83^{\circ}$ to $0.53^{\circ}$. Both measurements were sound and the inference from the first was not. Individual precision is the wrong criterion once three or more estimators are available, because a median can then reject rather than average, and rejecting well needs independnce. Both the corner angle and the hole rows are read off the plate's border, so a mislocalized border moves them together and a median of the two cannot detect it; the serial is found inside the frame's aperture and does not share that failure. A noisier third opinion that fails differently is worth more than a quieter one that fails the same way.

Pursuing that disagreement exposed a second fault, in a correction that had been earning its place under an earlier front end. The reported roll had been the corner angle plus a residual measured inside the rectified warp, added because the box fitted to a morphologically closed blob under-rotated at small tilts and left the missing roll as a rotation of the warp content. Against the present detector, which refines corners to sub-pixel and re-fits the edges to the image gradient, that residual is noise. At ground-truth rolls of $0$, $5$, $10$ and $15^{\circ}$ the corner angle alone errs by $0.28^{\circ}$, $0.61^{\circ}$, $0.39^{\circ}$ and $0.37^{\circ}$, while the sum errs by $0.39^{\circ}$, $0.64^{\circ}$, $0.59^{\circ}$ and $0.68^{\circ}$. The sum is worse at every value, including the non-zero rolls it existed to recover. Nothing had re-checked it when the front end changed. These two changes interact, so they are reported together: roll is $0.83^{\circ}$ with the residual and no serial, $0.72^{\circ}$ with neither, $0.53^{\circ}$ with both, and $0.44^{\circ}$ with the serial and no residual, the configuration reported throughout this paper.

Roll is recovered from three estimates, the tilts from one channel with the same redundancy. Four mounting holes form a rigid rectangle, so their imaged quadrilateral has its own vanishing points and the construction of Eq.~\eqref{eqn:vp} applies to them unchanged. Implemented and restricted to the frames where all four centers were measured, averaging that estimate with the outline's made both tilts $44\%$ worse: pitch moved from $6.82^{\circ}$ to $9.82^{\circ}$ and yaw from $8.36^{\circ}$ to $12.06^{\circ}$. Conditioning explains it, not principle. A vanishing point is the intersection of two nearly parallel lines, so its precision follows the length of the baselines and the accuracy of their endpoints, and the hole rectangle is smaller than the outline while its corners are circle-search centers accepted within $15$~pixels of a projected seed instead of sub-pixel-refined edges. This is a strictly worse estimator and not a second opinion of comparable quality, and averaging a good estimate with a worse one drags it. The same argument already excluded the perspective-$n$-point rotation; we record that it should have been applied here before the code was written, not after the sweep, since a construction being available on a feature does not make it accurate on that feature. Roll escapes the penalty because a hole row's inclination needs the two endpoints of one short segment instead of an intersection of two nearly parallel lines, so its aggregation survives where the tilts' does not.

Across the three angles the recovered attitude is not uniformly accurate, and we report the disparity rather than the average. In-plane roll comes back to $0.44^{\circ}$ of mean absolute error, with one qualification that must be stated because the comparison below depends on it: the sweep commands a level plate on every render, so that figure is the estimator's noise at zero roll and not its accuracy across the roll range, which the $0$ to $15^{\circ}$ diagnostic above puts at $0.28$ to $0.61^{\circ}$. The two out-of-plane tilts, scored across the angles the sweep does vary, come back to only $6.8^{\circ}$ and $8.4^{\circ}$, more than ten times worse, and the cause is neither the estimator nor the choice of estimator. Roll is an in-image quantity needing no perspective at all, whereas the two tilts must be read from the departure of the imaged quadrilateral from a scaled rectangle, and that departure is what a blob-fitted box removes, for the reason given in Sec.~\ref{sec:pose}. The limit is therefore corner localization and not geometry. Any pose method fed those corners fails identically, and a perspective-$n$-point solve on the same input measured about $5^{\circ}$. The gradient-based refinement recovers the perspective where the plate is large enough to support it a plate at a true $30^{\circ}$ comes back at $29.8^{\circ}$, with mean corner error falling from $15.4$ to $1.5$~pixels and the remaining gap is a detector limitation, addressable by the learned detector of Sec.~\ref{sec:conclusion} rather than by further work on the pose.

Because the sigmas are consumed downstream of the channels themselves, a weighting scheme can be evaluated by re-fusing distances already recorded, with no frames re-rendered and therefore no detector randomness entering the comparison. We used this to test the fixed size-derived priors of Eq.~\eqref{eqn:sizelaw} against two alternatives on the same $1224$ channel triples. Weighting each channel by per-frame measures of its quality left the fused error unchanged at $5.80\%$, and forcing the character channel to dominate raised it to $6.90\%$ while cutting the fraction of readings within $5\%$ from $72.0\%$ to $59.0\%$. This comparison is meaningful only if it fuses the way the pipeline fuses, and so applies the evidence gate of Sec.~\ref{sec:system} before weighting; an earlier version omitted it and reported $4.57\%$, scoring a system that was not the one under test. Two per-frame terms we tried and discarded are worth recording, both having been plausible. Scaling the character channel's variance by its glyph height in pixels cost $0.17\%$ points; the effect is real, that channel's bias running from $+4\%$ at $5$~m to $-43\%$ beyond $11$~m, but suppressing the channel further loses the small contribution it still makes at the size-derived weight. Dividing the hole channel's variance by the square root of the number of spans cost $0.32$ points and was simply wrong, the six spans being drawn from four hole centers and therefore not independent measurements that average down as $\sqrt{n}$.

The availability argument of Sec.~\ref{sec:system} is borne out by the sweep. That channel reported on $87.4\%$ of detected frames where the other two reported $100\%$, and on no frame was it the only one available. It earns its place as cover for the geometric channels failing together, not as the primary estimate.

\subsection{Operating Envelope and Consistency}

\begin{figure}[htbp]
\centering\includegraphics[width=\linewidth]{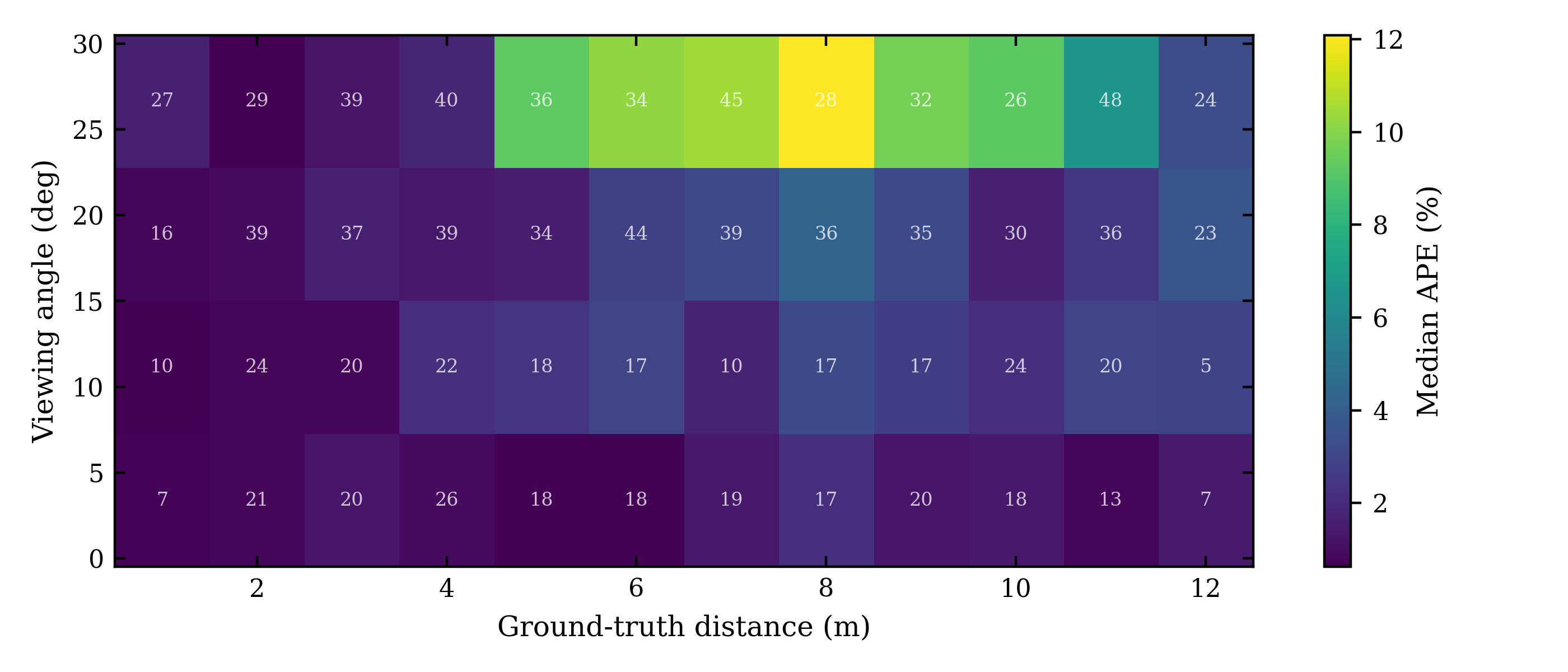}
\caption{Median fused error over the joint distance - angle envelope \label{Figure:heatmap}}
\end{figure}

Because a vehicle meets distance and angle jointly, we characterize the fused error over the two-dimensional envelope in Fig.~\ref{Figure:heatmap}, taking the median per cell, not the mean, since one gross detector failure would otherwise move a cell of about twenty-five samples by some fifteen points. The envelope is a plateau with one edge and not a gradient. Every one of the thirty-six cells below $30^{\circ}$ sits under $5\%$, averaging $1.9\%$, and that holds across the whole $1$ to $12$~m span, so distance on its own costs the fused estimate almost nothing. Degradation is confined to the $30^{\circ}$ row, and even there only past four meters. From $1$ to $4$~m that row runs $0.7$ to $2.0\%$, indistinguishable from frontal, then steps to between $9.4$ and $12.7\%$ from $5$~m before easing to $3.5\%$ at $12$~m. This step is the declined-correction population of Sec.~\ref{sec:results} seen from the other side, $30^{\circ}$ being where the foreshortening factor departs furthest from unity so that withholding the correction costs most, and the $+8.9\%$ bias measured on declined frames is the size of the step. A bound over the envelope is what a safety argument consumes, and this one is worth stating in two pieces: under $5\%$ below $30^{\circ}$ throughout, and up to $12.7\%$ in the oblique band beyond four meters.

A plate-based method must work not for one plate but for the many designs on the road. The fused error is below $8\%$ for fifty of the 51 registries, with a median across states of $3.7\%$ and the best near $2.7\%$, and the distribution is tight over that large majority. One design stands apart Alabama at $17.1\%$, whose plate face sits inside a bezelled surround so that its boundary is low-contrast and the detector localizes its corners imprecisely, biasing all three geometric channels by a common scale factor instead of failing outright. Delaware, previously the second outlier at $11\%$, now falls inside the band at $6.1\%$; its dark face on a dark mount is the same corner-localization problem, and the gradient-based edge refinement of Sec.~\ref{sec:pose} recovers it where a threshold could not. This is a detector-localization limitation on a small class of low-contrast designs, not a failure of the ranging principle, and it is the natural target of the further detector hardening noted in Sec.~\ref{sec:conclusion}. The consistency of the large majority follows directly from the two geometric channels, which do not depend on reading the plate and are therefore insensitive to the color scheme, header art, or serial format distinguishing one state's plate from another; the character channel varies more between designs, but the fusion's reliance on the geometric channels keeps the overall estimate stable. That the method generalizes across the jurisdictional variety without per-state tuning, outliers aside, is one of the more practically important findings.

\subsection{Uncertainty and Robustness}
\begin{figure}[htbp]
\centering\includegraphics[width=\textwidth]{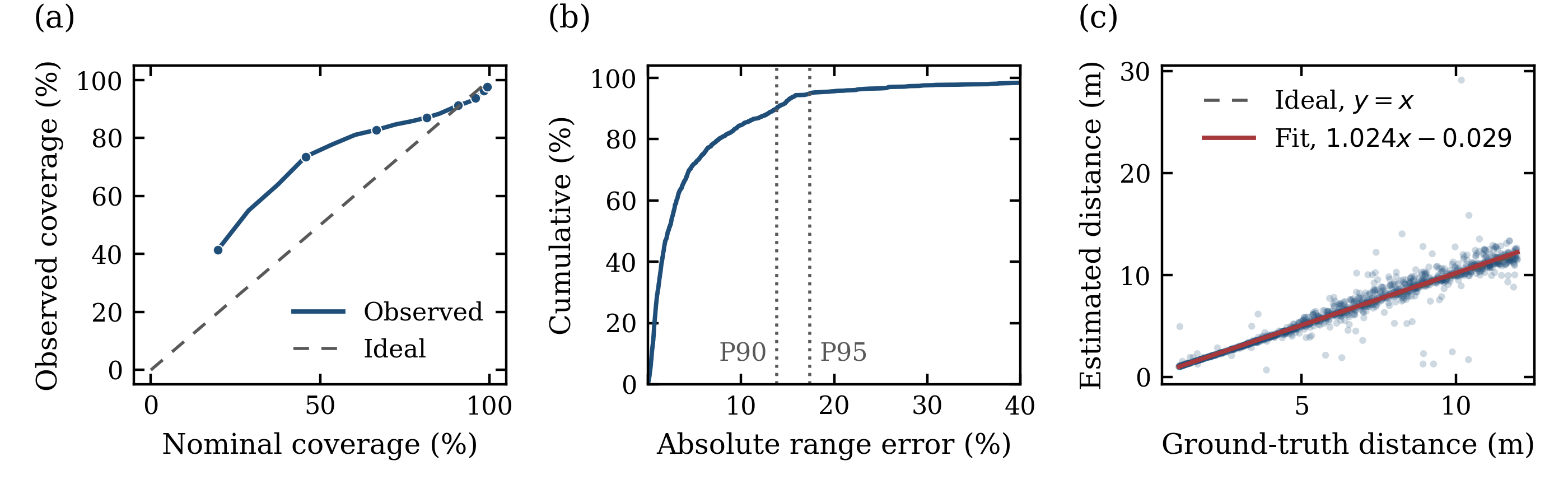}
\caption{Fused estimate, (a) the calibration of its reported uncertainty relative to the ideal coverage line, (b) The error distribution and its tail, and (c) Its agreement with truth in scale and offset\label{Figure:reliability}}
\end{figure}

An estimator reporting a variance makes a second claim alongside its distance, and a filter downstream acts on it, so we test it rather than assert it. Writing $z = (\hat{D}-D)/\hat{\sigma}$, a correctly calibrated estimator satisfies $\mathrm{ANEES} = E[z^2] = 1$ and places $68.3\%$ and $95.4\%$ of samples within one and two standard deviations. Over the continuous sweep the fusion attains $\mathrm{ANEES} = 4.78$ with $83.5\%$ and $93.4\%$ coverage, and the two statistics disagree in a way that is itself the result. ANEES is a mean of squared standardized errors and is therefore dominated by the few gross frames, while coverage is a count and describes the bulk, so a value far above unity beside a one-sigma coverage far above its nominal $68.3\%$ says the reported $\hat{\sigma}$ is conservative for ordinary frames and far too small for the rare catastrophic one. Reading either statistic alone would give the opposite verdict to the other, and the two conditions call for different remedies: a conservative core is a prior to retune, whereas an unmodelled tail is a detector failure no variance on the ranging channels was ever going to describe. Reaching this required deriving the priors from measurement instead of setting them by inspection. The relation $E|e| = \sigma\sqrt{2/\pi}$ makes the prior a channel's measured mean relative deviation divided by $0.798$, and with earlier hand-set figures of $0.06D$ and $0.08D$ the fusion was over-confident ($\mathrm{ANEES} = 1.67$, one-sigma coverage $61.4\%$) the unsafe direction, since a filter would then over-trust each reading. A prior derived this way is not fitted once and left alone, because it describes a channel's accuracy and the obliquity correction of Sec.~\ref{sec:pose} changed that accuracy on all three. Re-deriving gives $0.153D$, $0.054D$ and $0.088D$ for character, outline and hole channels earlier values are $0.28D$, $0.077D$ and $0.092D$. This refit was validated as the calibration gains were, a prior taken from the errors of a run and then scored on that same run being fitted to its own answer. Derived on one half of the sweep and scored on the held-out half, it improves the fused error from $4.60\%$ to $4.31\%$, and on the full sweep from $6.13\%$ to $5.79\%$. The error-propagation argument of Eq.~\eqref{eqn:sizelaw} predicts the ordering, and that holds through the refit unchanged the $65$~mm glyph still shows about three times the $305$~mm outline's relative uncertainty. Beyond the average error, a warning function must be specified on its tail behaviour. The fused estimate is within $5\%$ of truth on $72.0\%$ of frames and within $10\%$ on $84.6\%$, with $\mathrm{P}90$, $\mathrm{P}95$, and $\mathrm{P}99$ absolute errors of $13.8\%$, $17.4\%$, and $60\%$. Regressing the estimate on truth gives a slope of $1.024$ and an intercept of $-0.029$~m with $R^2 = 0.916$. The mean signed error over the sweep is $0.122$~m, so almost no net bias survives the per-channel calibration. Table~\ref{tab:statconst} collects every statistical constant in the pipeline with the measurement behind it, since a value chosen by eye is a result that cannot be reproduced. Figure~\ref{Figure:reliability} presents the three together: the coverage curve tracks the ideal line across its whole range, the error distribution reaches its knee near $10\%$, and the estimate versus truth cloud stays on the diagonal from $1$ to $12$~m without fanning out at range. The slight departure of the slope from unity paired with the positive intercept is a mild compression toward the middle of the range, not a scale error, and it displaces the fit from the identity line by less than $0.1$~m anywhere in the envelope.

Each subfigure in Fig. \ref{Figure:reliability} comes from one sample of the envelope, and the sweep is deterministic, so repeating the command reproduces them exactly and tells us nothing about how much of each is sampling variation. Drawing a second, independent sample separates the two, and the separation is sharp enough to change how the numbers should be read. Repeating the whole sweep at a different seed moves the fused mean from $5.79\%$ to $5.59\%$ and the median from $2.17\%$ to $2.34\%$; the fraction within $5\%$ moves by two parts in a thousand, the fraction within $10\%$ by three, the regression slope by $0.003$, and the one-sigma coverage by $1.4\%$ points. The bulk of the distribution is therefore stable to about $0.2\%$ point, and the second sample falls inside the confidence interval the first reported. The tail, however, is not stable at all. The $99$ percentile moves from $60.4\%$ to $68.3\%$, the largest single error from $18.9$ to $11.4$~m, and ANEES from $4.78$ to $6.81$ a swing of $42\%$ in a statistic quoted to three figures. This follows from what the tail is made of. A handful of gross detector failures dominates it, and the number sweep happens to draw is a property of sample, not of the method, so a tail statistic from a single run is good to roughly one significant figure and should be read that way. We therefore quote $\mathrm{P}99$ and $\mathrm{ANEES}$ to no more precision than that variation supports, and treat them as evidence that an unmodelled tail exists, not as measurements of its size. Across both samples the qualitative verdict survives and not its magnitude: both give a conservative core beside an unmodelled tail, and both put the fused mean near $5.7\%$. We report the two-sample comparison because a reproducible number and a repeatable one are different claims, and only the first is established by re-running a deterministic study.

\begin{table}[htbp]
\caption{Provenance of every statistical constant in the pipeline\label{tab:statconst}}
\centering\small
\begin{tabular}{@{}p{0.22\columnwidth}p{0.16\columnwidth}p{0.48\columnwidth}@{}}
\hline
Constant & Value & Provenance \\
\hline
Character prior & $0.153D$ & Pinhole error propagation from the $65$~mm feature, checked against $E|e| = \sigma\sqrt{2/\pi}$ \\
Outline prior & $0.054D$ & As above, from the $305$~mm feature \\
Hole-span prior & $0.088D$ & As above, from the $279.6$~mm feature \\
Consensus gate & $0.40$ & $1.4\sigma$ of ordinary inter-channel disagreement; swept from $0.05$ to none, total effect $0.41\%$ points \\
Hole gain & $0.9992 + 0.00742D$ & Fitted on half the sweep, scored on the held-out half ($7.18\%$ predicted, $7.63\%$ measured) \\
Measurement noise & $\mathrm{CV} = 0.06$ & The measured error of the fused input the filter consumes, not of one channel \\
Process noise & $\sigma_a = 4$~m/s$^2$ & A physical acceleration disturbance, about $0.4g$; sized from braking rather than tuned \\
Innovation gate & $5\sigma$, floor $20$~m/s & The floor is the fastest closing speed the $0.4$ - $13.2$~m envelope can present \\
Re-lock after & $6$ frames & The shortest streak that outlasts a transient mis-detection at $25$~Hz \\
Velocity deadband & $0.08$~m/s & Above frame-to-frame box jitter, below any approach speed that matters \\
Horizon penalty & $0.35$ & Score multiplier for a region above the horizon; reduced missed alerts from $11.6\%$ to $7.0\%$ \\
Warning thresholds & $1.5$~s, $2.4$~s & Taken from the NHTSA forward-collision specification, not chosen here \\
\hline
\end{tabular}
\end{table}

To probe conditions beyond clear daylight we ran the continuous driving scenario with overcast, dusk, night, rain, and glare applied to each frame after the plate was composited. Detection stayed complete across all five conditions, which is the first requirement of robustness. Ranging accuracy degraded gracefully and predictably with available light, the fused error staying near its clear-weather value under overcast and rising under night illumination, tracking the loss of edge contrast the character channel depends upon. Both geometric channels, relying on coarser structure than glyph edges, were the more stable under low light the same property that makes them the accurate channels at range makes them the resilient ones in the dark and the fusion again leaned on them automatically, which is the behavior the design intends.

\subsection{From Distance to Warning}\label{sec:fcw}
A ranging error matters only insofar as it changes what the vehicle is told, so the last experiment scores the output the application actually consumes. The continuous driving scenario runs for $883$ frames over ten successive targets, and at every frame the warning issued from the estimated range and closing speed is compared with the same decision rule returns when given the exact simulated distance and speed. Table~\ref{tab:fcw} reports the result, and it has to be read as an in-scene figure: unlike the pose study, the driving scenario places the plate on a vehicle body with the road and sky behind it, so it exercises plate localization in clutter as well as ranging.

Three numbers carry the meaning. The system agrees with the ground-truth warning on $85.7\%$ of frames. Of the $129$ frames on which the exact state of the world is danger the case a collision-warning function exists to catch $113$ raise a danger alert and a further $7$ raise a caution, so $7.0\%$ produce no alert at all; and of the $633$ quiet frames, $6.3\%$ raise an alert the ground truth does not support. These two rates are the ones a safety argument turns on, and they are asymmetric in consequence: a missed danger is a failure of the function, whereas a false alert is a failure of driver trust. Reporting them separately instead of as a single accuracy is what makes the trade-off visible, and reporting the missed alerts by severity is what stops a missed caution from being counted as though it were a missed danger. Time-to-collision, restricted to genuinely closing targets where it is a meaningful quantity, is recovered to a median of $0.27$~s. For a newly acquired target the warning becomes trustworthy after a median of $13$ frames, or $1.3$~s, the tracker needing several frames of range history before its closing-speed estimate supports any alert at all an acquisition latency a system specification must budget for and a single-frame accuracy figure does not reveal.

\begin{table}[htbp]
\caption{Forward-collision-warning agreement over the driving scenario\label{tab:fcw}}
\centering{%
\begin{tabular}{l r r r r}
\toprule
 & \multicolumn{3}{c}{Warning} & \\
\cmidrule(lr){2-4}
Ground-Truth & None & Caution & Danger & $n$ \\
\midrule
none & 593 & 19 & 21 & 633 \\
caution & 46 & 51 & 24 & 121 \\
danger & 9 & 7 & 113 & 129 \\
\bottomrule
\end{tabular}
}%
\end{table}

The in-scene fused error over the same run has a median of $2.8\%$, close to the controlled result, but a mean of $13.1\%$, and the gap between the two is the far-range localization failure of Sec.~\ref{sec:conclusion}, not a degradation of the ranging channels. This failure also limits the warning agreement, since the missed dangers concentrate where the plate subtends fewest pixels and the detector is most easily drawn to a competing structure. The temporal filter recovers part of it, reducing the fraction of grossly wrong frames from $9.1\%$ to $8.0\%$ here and by considerably more in the adverse-condition run, which is the behavior the innovation gate exists to provide. We report these figures as the end-to-end performance of the present implementation in a synthetic scene, and distinguish them throughout from the controlled per-channel accuracy of Sec.~\ref{sec:results}, which characterizes the ranging method itself; the two measure different things and should not be quoted interchangeably.

Finally we isolate what each improvement contributes, the aggregate numbers above already including all three. Robust localization is responsible for the completeness of detection: the frames that would previously have been lost to a body lock or a character-cluster collapse are the low-contrast and sparse-plate cases, and converting them into correct detections is the precondition for every channel. The consensus gate is what allows the fusion to reach the accuracy of the best single channel without inheriting its worst case; in its absence, a corrupted channel that happened to agree with a second corrupted channel could pull the estimate away from truth through their small prior variances, and the median-based rejection forecloses that. The innovation gate removes the isolated large errors a single misdetection would otherwise inject into the filtered range and time-to-collision. Its contribution is not visible in a per-frame average. It shows in the absence of the transient spikes that separate a stable track from an occasional false warning.

\subsection{Design Guidance}\label{sec:discuss}

The characterization supports a small number of concrete recommendations, and it also clarifies where the method belongs in a vehicle's sensor suite.

The first recommendation runs counter to a natural intuition: prefer the coarse geometric features over the fine one. Tempting though it is to assume that the most information-rich feature the crisply printed character, which the eye reads most easily makes the best ruler, the error-propagation relation and the results agree that it does not. Both the plate outline and the hole span are more accurate and more robust than character height across almost the whole envelope, and a minimal implementation using only the outline would already reach $4.32\%$ with no character reading at all. Character height should be treated as a close-range refinement the fusion is free to down-weight, not as the primary channel.

The second is that fusion is worth adopting for robustness first and accuracy second. Once the three channels are calibrated the fused error does edge below the best single channel, but the margin is a fraction of $1\%$ and would not on its own justify the added complexity. What justifies it is the fusion turning three fallible channels into one self-checking estimate whose worst case is bounded by the agreement of the others rather than by any single feature. Safety systems value that far above a fractional gain in accuracy. It is the difference between an estimate that fails visibly and one that fails silently.

The third concerns how the accuracy should be specified. A single global figure understates the method close and overstates it far, so the distance characterization gives the real budget: under $0.1$~m near, single-digit-percent through the mid-range, and a graceful rise at the far edge carried there by the geometric channels. Because the joint envelope is smooth, this budget can be stated as a bound over an operating region, as a functional-safety case requires.

These recommendations locate plate-based ranging within a layered perception stack~\cite{janai2020computer,van2018autonomous,kukkala2018advanced}. Plate-based ranging is a passive, geometry-based channel whose accuracy is now known as a function of the operating conditions, and whose failure modes loss of the plate to occlusion or absence, extreme obliquity, and very low plate-to-background contrast are disjoint from the failure modes of radar, lidar, and stereo. This disjointness is the point. The method is proposed not as a replacement for an active rangefinder but as an inexpensive, independent corroborator whose agreement raises confidence and whose disagreement is itself informative, in the manner redundant perception design calls for.

\section{Conclusion}\label{sec:conclusion}

We presented a license-plate-based monocular ranging system and validated it in a metrology-exact simulator. With strengthened plate localization on low-contrast and cluttered frames, fused the character, outline, and hole-span channels by a consensus-gated inverse-variance rule that discards a corrupted channel before it biases the estimate, and protected the temporal filter from mis-detections with an innovative gate. A simulator projecting standardized plates by the same pinhole geometry the method inverts supplied exact ground truth without a road trial, across all 51 USA jurisdictions, distances to $12$~m, viewing angles to $30^{\circ}$, and five illumination conditions. The plate is detected in every frame. The plate-outline channel reaches $4.32\%$ mean absolute percentage error, and the calibrated fusion returns $5.79\%$ mean and $2.17\%$ median without improving on that channel. In-plane roll comes back to within one degree and the two out-of-plane tilts to within several. The estimate holds below $8\%$ for 50 of the 51 jurisdictions and degrades gracefully with distance, angle, and light.

Three limitations bound these results, and all three fall outside the ranging mathematics. Ground truth is exact because the scene is synthesized, so the figures are a geometry-limited result and instrumented on-road measurement at surveyed distances is the natural extension. Accuracy is measured in a controlled background, and in a full driving scene the detector intermittently locks onto a wrong region at long range, where the plate subtends fewest pixels; the three channels then measure that region in agreement, so the consensus gate cannot reject them. Gross failures reach roughly a third of frames beyond $11$~m in that setting, though median error is largely unchanged, and the innovation gate catches about two in five, cutting them from $9.7\%$ to $5.9\%$ over a $900$-frame scenario. Within the fusion, the outline and hole-span errors correlate at $+0.928$ because the circle search is seeded from a pose fitted to the outline corners, which understates the fused standard deviation by about $39\%$ through Eq.~\eqref{eqn:corr}; the distance is unaffected, but a full covariance is needed before that uncertainty is passed downstream. Runtime is $0.4$ to $2.4$~s per frame in this Python prototype, of which plate detection is roughly $99\%$ and the pose solve and character measurement together are about $23$~ms. The ranging is therefore essentially free, and every remaining gap is a detector problem.

Robust localization, consensus-gated fusion, and innovation-gated filtering together convert a plate-ranging pipeline that worked on clean plates into one that holds across a full operating envelope, with complete detection and a self-checking distance estimate.
The coarse geometric features of the plate, its outline and its mounting-hole span, are the more accurate and the more robust ranging channels, and the fine character height is best used as a close-range refinement; fusion is warranted by robustness and self-checking rather than by any accuracy gain,  since it matches the best single channel.
A metrology-exact testbed such as MetroSim validates the system to a precision a road test cannot easily match, exercising the full jurisdiction, distance, angle, and illumination envelope reproducibly and retargeting to a new camera or vehicle class by a
one-line specification change.

\section*{Acknowledgments}
This research was partially supported from the University of Michigan Office of the Vice President for Research through Bold Challenges and AIIM Seed Networking Award.

\section*{Data and Code Availability}
The data and code will be made publicly available once the paper is accepted.

\section*{Nomenclature}

\begin{description}[leftmargin=3.2em,style=sameline,labelwidth=2.6em]
\item[$D$] true camera-to-plate distance (m)
\item[$\hat{D}$] fused distance estimate (m)
\item[$d_i$] distance reported by channel $i$ (m)
\item[$f$] focal length at the working image width (pixels)
\item[$H$] true (physical) height of an imaged feature (m)
\item[$H_s$] regulated reference character height (m)
\item[$h$] imaged height of a feature (pixels)
\item[$S$] Kalman innovation covariance (m$^2$)
\item[$s_i^2$] prior variance of channel $i$ (m$^2$)
\item[$\tilde{s}_i^2$] disagreement-inflated variance of channel $i$ (m$^2$)
\item[$W$] image width (pixels)
\item[$w_i$] inverse-variance weight of channel $i$
\item[$\theta_H$] horizontal field of view (deg)
\end{description}

\subsection*{Abbreviations}

\begin{description}[leftmargin=3.2em,style=sameline,labelwidth=2.6em]
\item[AAMVA] American Association of Motor Vehicle Administrators
\item[ANEES] average normalized estimation error squared
\item[ANSI] American National Standards Institute
\item[CV] coefficient of variation
\item[FHWA] Federal Highway Administration
\item[FIPS] Federal Information Processing Standard
\item[MAPE] mean absolute percentage error
\item[NHTSA] National Highway Traffic Safety Administration
\item[PnP] perspective-$n$-point
\item[RANSAC] random sample consensus
\end{description}

\bibliographystyle{unsrtnat}
\bibliography{JAVS_arxiv}

\end{document}